\documentclass[a4paper,fleqn]{cas-dc}
\usepackage[numbers,sort&compress]{natbib}
\usepackage{amsmath,amssymb,amsthm,booktabs,tabularx,graphicx}
\usepackage{adjustbox,xurl}
\usepackage{xcolor}
\definecolor{bestvalue}{HTML}{A51C30}
\definecolor{secondvalue}{HTML}{1565C0}
\newcommand{\firstmean}[1]{\textcolor{bestvalue}{#1}}
\newcommand{\secondmean}[1]{\textcolor{secondvalue}{#1}}
\usepackage{algorithm,algpseudocode}
\usepackage[section]{placeins}
\hypersetup{hypertexnames=false,hyperfootnotes=false}
\usepackage{kbs-layout}
\newcommand{\tablefont}{\rmfamily\fontsize{7.2}{8.6}\selectfont}

\newcommand{\R}{\mathbb{R}}
\newcommand{\E}{\mathbb{E}}
\newcommand{\diag}{\operatorname{diag}}
\newcommand{\tr}{\operatorname{tr}}
\newcommand{\sg}{\operatorname{sg}}
\newcommand{\softplus}{\operatorname{softplus}}
\newcommand{\ours}{FaCoDiff}
\newsavebox{\bestnormal}
\newsavebox{\bestbold}
\newcommand{\best}[1]{\begingroup
  \sbox{\bestnormal}{$#1$}\sbox{\bestbold}{\textcolor{bestvalue}{$\mathbf{#1}$}}%
  \resizebox{\wd\bestnormal}{\ht\bestnormal}{\usebox{\bestbold}}%
  \endgroup}
\begin{document}
\shorttitle{FaCoDiff for multivariate probabilistic forecasting}
\shortauthors{J. Li et al.}
\title[mode=title]{FaCoDiff: Factor-Correlated Diffusion for Multivariate Probabilistic Forecasting}
\author[1]{Jiaxuan Li}
\ead{lijiaxuan@std.uestc.edu.cn}
\author[2]{Yao Liu}
\cormark[1]
\ead{liuyao@uestc.edu.cn}
\author[1]{Tingting Dai}
\ead{ttdai@std.uestc.edu.cn}
\author[1]{Yuewei Xue}
\ead{202511081620@std.uestc.edu.cn}
\author[3]{Shijie Dong}
\ead{tou@ruri.waseda.jp}
\author[4]{Zixiang Jin}
\ead{handsomejzx@163.com}
\author[1]{Xiaoyu Guo}
\ead{202611080310@std.uestc.edu.cn}
\author[1]{Qiao Liu}
\ead{qliu@uestc.edu.cn}

\affiliation[1]{
  organization={School of Computer Science and Engineering,
    University of Electronic Science and Technology of China},
  city={Chengdu},
  postcode={611731},
  country={China}
}
\affiliation[2]{
  organization={School of Information and Software Engineering,
    University of Electronic Science and Technology of China},
  city={Chengdu},
  postcode={610054},
  country={China}
}
\affiliation[3]{
  organization={Graduate School of Information, Production and Systems,
    Waseda University},
  city={Tokyo},
  postcode={169-8050},
  country={Japan}
}
\affiliation[4]{
  organization={School of Economics and Management,
    Dalian University of Technology},
  city={Dalian},
  postcode={116024},
  country={China}
}
\cortext[1]{Corresponding author: Yao Liu.}

\begin{abstract}
Multivariate probabilistic forecasting requires modeling cross-variable
dependence alongside marginal uncertainty. The challenge is to learn this
dependence while controlling marginal noise scales and avoiding dense
covariance costs. We propose Factor-Correlated Diffusion (FaCoDiff), which
couples a compact conditional statistical estimator with structured
denoising. History-conditioned factor loadings use distinct coefficients
for forecast-center correction, scale adjustment and noise dependence.
Diagonal compensation preserves the estimated marginal noise variances,
while low-rank operators enable residual Gaussian likelihood evaluation
and noise transport without dense covariance storage or factorization.
Factor-derived feature modulation explicitly conditions the denoiser,
and a geometry-aware objective accounts for errors after noise transport.
The same conditional geometry governs forward perturbations and reverse
sampling.
Across nine datasets, FaCoDiff achieves the lowest mean variogram score
on eight datasets and ranks among the top two on 43 of 54 dataset--metric
combinations in the reported comparisons. Matched ablations on Solar and
Weather show that factor covariance improves mean Energy and Variogram
scores under all four combinations of center and scale correction.
Disrupting cross-variable sample pairings while preserving marginal
trajectories worsens joint and aggregate scores on both datasets, demonstrating the
predictive value of generated cross-variable dependence.
\end{abstract}

\let\printorcid\relax
\maketitle
\hypersetup{pdfauthor={Jiaxuan Li; Yao Liu; Tingting Dai; Yuewei Xue; Shijie Dong; Zixiang Jin; Xiaoyu Guo; Qiao Liu}}

\section{Introduction}
\label{sec:introduction}
Multivariate forecasting requires more than accurate predictions for each
variable in isolation. In energy and transportation systems, simultaneous
deviations across variables determine how uncertainty accumulates across
locations and services. A probabilistic forecaster must therefore describe
both the uncertainty of individual outcomes and the dependence among them.
Marginal scores assess uncertainty for individual variables, whereas
multivariate scores also assess how generated outcomes combine
\cite{gneiting2007scoring,scheuerer2015variogram}. A model can therefore
improve marginal forecasts without recovering useful cross-variable
co-movement; both properties require evaluation.

Diffusion models offer a flexible way to generate conditional forecasts
through iterative denoising \cite{ho2020ddpm,ye2025nsdiff}.
Data-dependent Gaussian reference distributions connect this generation
process to conditional statistical estimation
\cite{lee2022priorgrad,yang2026cwgen}. In time-series forecasting, NsDiff
adapts the reference center and marginal scales to the history
\cite{ye2025nsdiff}, while CW-Gen incorporates conditional covariance
through whitening \cite{yang2026cwgen}. These developments motivate a
specific modeling question: how can conditional centers, marginal scales
and cross-variable noise dependence be learned together without the
storage and factorization costs of dense covariance estimation?

Two issues arise when constructing such a model. First, diagonal noise
provides variable-specific scales but does not explicitly encode shared
residual fluctuations. Dense covariance expresses these fluctuations
at the cost of quadratic storage and dense matrix factorizations.
Low-rank covariance is an established alternative in probabilistic
forecasting \cite{salinas2019copula,zheng2024correlated}; the question here
is how to use it in a conditional estimator and a diffusion generator
without losing control of the marginal noise scales. Second, adding a
positive-semidefinite low-rank term to a prescribed diagonal also increases its
marginal variances. Dependence should therefore be introduced with
diagonal compensation, so that changes in noise correlation are not
automatically changes in the estimated marginal variances.

We introduce \ours{}, a factor-correlated diffusion model that addresses
these issues through a compact conditional estimator. Base predictors
provide a forecast center and residual variance scales. A
history-conditioned loading representation then supports separate
coefficients for center adjustment, marginal scale adjustment and noise
dependence. Shared loadings provide a compact statistical representation;
distinct coefficients allow center adjustments and stochastic fluctuations
to differ in magnitude. Factor-correlation normalization
\cite{murray2013copula,salinas2019copula} compensates the diagonal when
dependence is introduced, preserving the estimated marginal noise
variances. A residual Gaussian score and center supervision train this
estimator, while a denoiser learns under the corresponding structured
perturbations. Factor-derived features explicitly modulate the denoiser,
and its objective balances white-noise error with the error after
covariance transport. The same conditional geometry is used for generation, with
likelihood and noise transformations computed directly from diagonal and
low-rank factors. This design assigns complementary roles to residual
estimation and denoising: the estimator learns a structured statistical
reference, while the reverse chain learns a predictive distribution
beyond direct Gaussian sampling.

Our contributions are threefold:
\begin{itemize}
 \item We propose FaCoDiff, a compact conditional estimator and diffusion
 generator for multivariate probabilistic forecasting. History-conditioned
 loadings and distinct coefficients coordinate center corrections, marginal
 scales and residual dependence, while diagonal compensation introduces
 noise correlation at the estimated marginal variances.
 \item We develop geometry-conditioned denoising and a low-rank learning
 and sampling implementation. Factor-derived feature modulation exposes
 the conditional dependence to the denoiser, whose objective balances
 white-coordinate and trace-normalized transported errors. Covariance computations operate on
 diagonal and low-rank factors with $r\times r$ matrix decompositions,
 avoiding dense covariance storage and factorization.
 \item We evaluate on nine datasets and test dependence separately from
 marginal corrections through matched Solar and Weather ablations, controlled synthetic
 data and sample-pairing interventions. Direct Gaussian controls and
 complete-generation measurements characterize the predictive value and
 computational cost of diffusion generation.
\end{itemize}

\section{Related work}
\label{sec:related-work}
\subsection{Multivariate probabilistic forecasting}
Probabilistic forecasting represents conditional uncertainty
\cite{lim2021survey} through recurrent likelihood models and neural state-space
models \cite{salinas2020deepar,rangapuram2018deepstate}.
Quantile regression \cite{koenker1978quantiles} and multi-horizon neural
forecasters \cite{wen2017mqrnn,lim2021tft} avoid a full likelihood, but
marginal quantiles alone do not determine multivariate dependence.

Deep Factors combines global neural factors with local probabilistic components
\cite{wang2019deepfactors}. Multivariate Gaussian recurrent models and
Gaussian copula forecasters use low-rank covariance, with the latter
additionally coupling marginal transformations to the dependence model
\cite{salinas2019copula}. Correlated-error models combine
diagonal-plus-low-rank contemporaneous covariance with temporal error
structure \cite{zheng2024correlated}. Conditioned normalizing flows instead
represent multivariate densities through invertible transformations
\cite{dinh2017realnvp,rasul2021flows}. In \ours{}, factors parameterize
conditional residual covariance and diffusion noise; the learned
reverse process determines the predictive distribution.

\subsection{Conditional diffusion forecasting}
Diffusion reverses stochastic perturbations
\cite{sohldickstein2015diffusion,ho2020ddpm,song2021sde}, with choices in
conditioning and sampling \cite{chang2026fundamentals}. Time-series models
use historical states for autoregressive generation \cite{rasul2021timegrad}
or observation masks for partially observed inputs
\cite{tashiro2021csdi}. Other strategies guide
generation through future mixup and autoregressive initialization
\cite{shen2023timediff}, observation-based self-guidance
\cite{kollovieh2023tsdiff}, or pretrained representations
\cite{yang2027gpd4ts}. Their generative networks can model dependence even
with independent perturbation noise.

Other methods condition the reference distribution. PriorGrad adapts a
Gaussian prior \cite{lee2022priorgrad}; CARD centers diffusion on a mean
estimate \cite{han2022card}. TMDM incorporates Transformer forecasts into
forward and reverse processes \cite{li2024tmdm}. NsDiff adds conditional
marginal variance and an uncertainty-aware schedule \cite{ye2025nsdiff}.
DiffPTS derives a Gaussian likelihood objective for statistical estimators
from the location--scale diffusion ELBO \cite{ye2026diffpts}. CN-Diff instead
combines nonlinear data transformations with learnable forward conditions
\cite{rishi2025cndiff}.

CW-Gen estimates conditional means and dense sliding-window covariance
targets, using whitening to connect them to diffusion and flow generation
\cite{yang2026cwgen}. FaCoDiff instead learns compact conditional factors
from forecast residuals and uses diagonal compensation to control the
marginal noise variances. Both approaches use a conditional Gaussian
reference; FaCoDiff couples its residual-based estimator to low-rank
likelihood and noise-transport operators.

\subsection{Marginal control and scalable covariance}
Marginal adaptation and shared uncertainty are distinct tasks. Instance
normalization and data-dependent attention address non-stationarity
in time-series representations \cite{kim2022revin,liu2022nonstationary}; variable-attention
and dynamic-graph encoders model interactions in the observed history
\cite{liu2024itransformer,hou2025pmedgn}. Noise covariance instead describes
fluctuations around a forecast center. Dynamic conditional correlation
separates variances from correlations \cite{engle2002dcc}; Gaussian copula
factor models normalize loadings to obtain unit-diagonal correlations
\cite{murray2013copula}. \ours{} uses this normalization to compensate the
diagonal increase caused by adding a low-rank term, preserving the estimated
marginal noise variances.

Shrinkage stabilizes covariance estimates \cite{ledoit2004shrinkage}, while
diagonal-plus-low-rank representations reduce storage and factorization costs
\cite{salinas2019copula,zheng2024correlated}. Compared with CW-Gen's dense
estimator and window-based covariance targets, \ours{} learns conditional
factors from forecast-residual Gaussian scores. Shared loadings support
distinct center, scale and noise-strength coefficients, and standard
low-rank matrix identities \cite{golub2013matrix} permit likelihood
evaluation and noise transport without dense covariance matrices.
This is complementary to reducing denoising evaluations through fast samplers
\cite{song2021ddim,lu2022dpmsolver}.

\section{Method}
\label{sec:method}
\subsection{Problem formulation and overview}
Given a history $X\in\R^{N\times d}$, probabilistic forecasting learns
$p(Y\mid X)$ for a future $Y\in\R^{H\times d}$. Here $N$, $H$, and $d$
denote the history length, forecast horizon, and number of variables.
We write $Y_0=Y$, use $h$ for forecast steps and $t$ for diffusion steps,
and treat horizon-specific vectors such as $Y_{t,h}$ and $\mu_h$ as
columns. Batch indices are omitted except in the training objectives.

\begin{figure*}[pos=tbp]
\centering
\includegraphics[width=\textwidth]{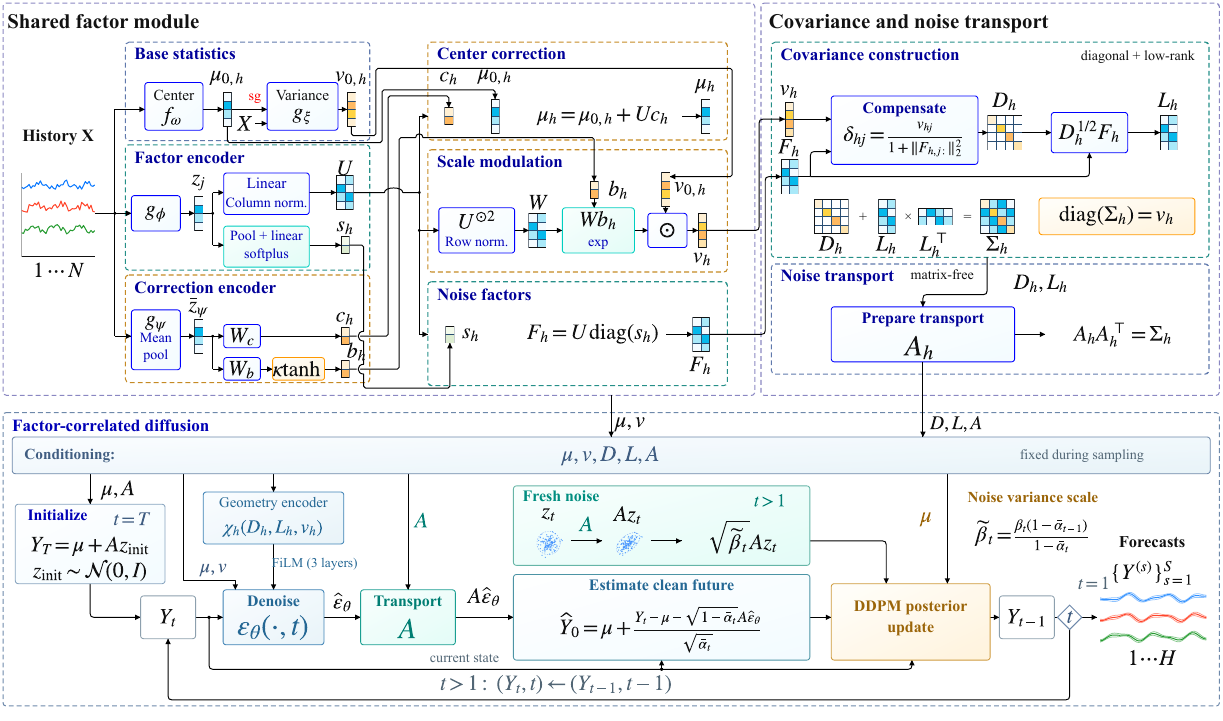}
\caption{Architecture of FaCoDiff. Shared loadings support center correction, scale modulation and correlated noise. Diagonal compensation preserves marginal noise variances. Factor-derived geometry modulates the denoiser, and the transport $A$ is reused throughout reverse generation.}
\label{fig:architecture}
\end{figure*}

\ours{} first estimates conditional statistics and then generates
forecasts by structured denoising (Figure~\ref{fig:architecture}). For each
history, its statistical estimator produces a center
$\mu$, positive marginal noise variances $v$, and an implicit covariance
representation $\{\Sigma_h\}_{h=1}^H$:
\begin{equation}
 X\longmapsto(\mu,v,\{\Sigma_h\}_{h=1}^H)
 \longmapsto\{Y^{(s)}\}_{s=1}^S.
\label{eq:overview}
\end{equation}
The center and scale corrections refine the statistical reference, while
the factor covariance represents shared residual fluctuations. Their
coefficients use common loadings but are estimated separately. Diagonal
compensation controls the noise variances when these correlations are
introduced. The estimator learns from residual likelihood and center
supervision; the denoiser learns to reverse perturbations with this
geometry through explicit factor-derived conditioning and a geometry-aware
denoising objective. Low-rank operators implement the likelihood and both diffusion
directions without constructing a dense covariance matrix.

\subsection{Base center and variance prediction}
\label{sec:base-predictors}
We first estimate variable-specific centers and residual variance scales:
\begin{equation}
\begin{aligned}
 \mu_0(X)&=f_\omega(X)\in\R^{H\times d},\\
 v_0(X)&=g_\xi(X,\sg(\mu_0(X)))\in\R_{>0}^{H\times d}.
\end{aligned}
\label{eq:base-predictors}
\end{equation}
The center predictor is a Non-stationary Transformer
\cite{liu2022nonstationary}, with available calendar features included in its
inputs. The stop-gradient operator $\sg$ prevents the variance predictor
from changing the center through its input features.

The variance predictor uses local historical activity, observed increments,
and the anticipated center trajectory. Define
$\rho(x)=\operatorname{sign}(x)\log(1+|x|)$ and
\begin{equation}
\begin{aligned}
 m_{nj}&=\frac1w\sum_{\ell=0}^{w-1}X_{n-\ell,j},\\
 a_{nj}&=\log\!\left(1+\left[
 \frac1w\sum_{\ell=0}^{w-1}X_{n-\ell,j}^2-m_{nj}^2
 \right]_+\right),\\
 q_{nj}&=\rho(X_{nj}-X_{n-1,j}),\\
 p_{hj}&=\rho(\sg(\mu_{0,hj})-X_{Nj}).
\end{aligned}
\label{eq:variance-features}
\end{equation}
Here $j$ indexes variables, $n$ indexes history positions, $w=24$, and
$[x]_+=\max(x,0)$. Left boundaries repeat the first observation, so
$q_{1j}=0$. A variable-shared MLP maps these features to
\begin{equation}
 v_{0,:,j}=\softplus\!\left(
 \operatorname{MLP}_{\xi}([a_{:,j};q_{:,j};p_{:,j}])\right)+\eta.
\label{eq:mean-conditioned-variance}
\end{equation}
It has two ReLU hidden layers of width 512; $\eta=10^{-6}$ ensures positive
outputs. Both predictors learn from forecast residuals through the
objectives in Section~\ref{sec:training}.

\subsection{Conditional factor estimation}
\label{sec:estimator}
The loading matrix $U(X)\in\R^{d\times r}$ represents shared variation.
Separate coefficients adjust centers and scales, while horizon-specific
strengths $s_h(X)\in\R_{>0}^r$ determine noise factors.

\paragraph{History-conditioned loadings}
A temporal MLP $g_\phi$, shared across variables, encodes each history.
Its individual and pooled features determine the loadings and strengths:
\begin{equation}
\begin{aligned}
 z_j&=g_\phi(X_{:,j}),\qquad \bar z=\frac1d\sum_jz_j,\\[2pt]
 \widetilde U_{j,:}^{\top}&=W_Uz_j+a_U,\\
 U_{:,k}&=\frac{\widetilde U_{:,k}}
 {\max(\|\widetilde U_{:,k}\|_2,\varepsilon)},\\[2pt]
 s_{1:H}&=\operatorname{reshape}_{H\times r}
 [\softplus(W_s\bar z+a_s)].
\end{aligned}
\label{eq:factorhead}
\end{equation}
The encoder features are column vectors, $s_{1:H}$ stacks the rows
$s_h^\top$, and $\varepsilon=10^{-8}$ is a numerical floor.
The loadings change with history but are shared across forecast steps;
$s_h$ adjusts their contribution at each step. Column normalization fixes
their scale while allowing nonorthogonal directions. The factor-head
parameter count is independent of $d$, and $r$ specifies its capacity.

\paragraph{Center and scale corrections}
The center and scale corrections share $U$ with the noise factors but
have their own coefficients. A second variable-shared encoder $g_\psi$
produces these coefficients from pooled history features:
\begin{equation}
\begin{aligned}
 \bar z_\psi&=\frac1d\sum_jg_\psi(X_{:,j}),\\[2pt]
 C_\mu&=\operatorname{reshape}_{H\times r}(W_c\bar z_\psi+a_c),\\
 C_v&=\kappa\tanh\!\left[
 \operatorname{reshape}_{H\times r}(W_b\bar z_\psi+a_b)\right].
\end{aligned}
\label{eq:correction-heads}
\end{equation}
Let $c_h=(C_\mu)_{h,:}^{\top}\in\R^r$ and
$b_h=(C_v)_{h,:}^{\top}\in\R^r$. The corrected statistics are
\begin{equation}
 \mu_h=\mu_{0,h}+Uc_h,
\label{eq:location}
\end{equation}
\begin{equation}
\begin{aligned}
 W_{jk}&=\frac{U_{jk}^2}{\max(\sum_\ell U_{j\ell}^2,\varepsilon)},\\
 v_{hj}&=\max\!\left\{v_{0,hj}\exp\!\left(\sum_kW_{jk}b_{hk}\right),\eta\right\}.
\end{aligned}
\label{eq:scale}
\end{equation}
Thus the center correction lies in the loading span, whereas squared
loadings weight factor-specific log-variance adjustments. With $\kappa=0.7$,
the variance multiplier is bounded by $[e^{-\kappa},e^\kappa]$.
Zero-initialized coefficient heads recover $(\mu_0,v_0)$.

\paragraph{Diagonal compensation}
Adding a low-rank term directly to $\diag(v_h)$ would also inflate the
diagonal. We compensate this contribution using
$F_h=U\diag(s_h)$, whose $j$th row is $f_{hj}$:
\begin{equation}
\begin{alignedat}{2}
 \delta_{hj}&=\frac{v_{hj}}{1+\|f_{hj}\|_2^2},\qquad &D_h&=\diag(\delta_h),\\
 L_h&=D_h^{1/2}F_h, &\Sigma_h&=D_h+L_hL_h^\top.
\end{alignedat}
\label{eq:covariance}
\end{equation}
The $j$th diagonal entry is
$\delta_{hj}(1+\|f_{hj}\|_2^2)=v_{hj}$, so
$\diag(\Sigma_h)=v_h$. This is a factor-correlation normalization
\cite{murray2013copula,salinas2019copula}.
Positive definiteness follows because, for any nonzero $a$,
$a^\top\Sigma_ha=a^\top D_ha+\|L_h^\top a\|_2^2>0$.
The factors $L_h$ combine $U$, $s_h$ and $D_h^{1/2}$ to control dependence
at fixed $v_h$.

\subsection{Factor-correlated diffusion}
\label{sec:diffusion}
For a given history, $(\mu_h,\Sigma_h)$ remain fixed along the diffusion
chain. Let $A_hA_h^\top=\Sigma_h$ define an invertible noise transport.
The forward and reverse processes use this same transport, evaluated
without forming a dense matrix as described in Section~\ref{sec:lowrank}.

\paragraph{Forward perturbation}
Given a schedule $\beta_t\in(0,1)$, set $\alpha_t=1-\beta_t$,
$\bar\alpha_t=\prod_{\tau=1}^t\alpha_\tau$, and $\bar\alpha_0=1$. Define
\begin{equation}
 q(Y_{t,h}\mid Y_{t-1,h},X)=\mathcal N\!\left(
 \mu_h+\sqrt{\alpha_t}(Y_{t-1,h}-\mu_h),\,
 \beta_t\Sigma_h\right).
\label{eq:forward-kernel}
\end{equation}
Training samples an arbitrary diffusion step directly through
\begin{equation}
 Y_{t,h}=\mu_h+\sqrt{\bar\alpha_t}(Y_{0,h}-\mu_h)
 +\sqrt{1-\bar\alpha_t}A_h\epsilon_h,
 \quad\epsilon_h\sim\mathcal N(0,I_d).
\label{eq:forward}
\end{equation}
The perturbations are independent across $h$; $\Sigma_h$ specifies
cross-variable noise dependence within each forecast step. Together with
horizon-wise denoising, this yields a predictive distribution that
factorizes across forecast steps conditional on $X$.

\paragraph{Geometry-conditioned denoising}
The denoiser predicts pre-transport noise from the noisy future and time
step, conditioned on the center, marginal variance and covariance factors.
To expose dependence independently of
the marginal scales, define $V_h=\diag(v_h)$ and the geometry features
\begin{equation}
 \chi_h=\left[\log(\delta_h\oslash v_h);\,
 \sqrt d\,\operatorname{vec}(V_h^{-1/2}L_h)\right],
\label{eq:geometry-features}
\end{equation}
where $\oslash$ denotes elementwise division. A two-layer MLP maps
$\chi_h$ to three pairs of feature-wise modulation vectors
$(a_h^{(\ell)},b_h^{(\ell)})$, $\ell=1,2,3$.
Starting with $u_h^{(0)}=[Y_{t,h};\mu_h;v_h]$, the hidden updates are
\begin{equation}
 u_h^{(\ell)}=\softplus\!\left[
 \left(1+\tfrac12\tanh a_h^{(\ell)}\right)\odot
 T_{\theta,\ell}(u_h^{(\ell-1)},t)+b_h^{(\ell)}\right],
\label{eq:geometry-film}
\end{equation}
where $T_{\theta,\ell}$ is a linear projection multiplicatively modulated
by a learned diffusion-step embedding \cite{ye2025nsdiff}. A final linear
projection produces $\widehat\epsilon_{\theta,h}$. Weights are shared
across forecast steps. The geometry MLP's output layer is initialized to
zero, so its modulation initially leaves the time-conditioned hidden
projections unchanged. Geometry modulations are prepared at the start of
each sample chunk and reused across its reverse steps. Inputs
remain in data coordinates; $A_h\widehat\epsilon_{\theta,h}$ gives the
correlated noise estimate.

\paragraph{Reverse generation}

Sampling starts from the conditional Gaussian reference by drawing
$Y_{T,h}=\mu_h+A_hz_{\mathrm{init},h}$ with
$z_{\mathrm{init},h}\sim\mathcal N(0,I_d)$. At each reverse step, we first
reconstruct
\begin{equation}
 \widehat Y_{0,h}=\mu_h+
 \frac{Y_{t,h}-\mu_h-\sqrt{1-\bar\alpha_t}A_h\widehat\epsilon_{\theta,h}}
 {\sqrt{\bar\alpha_t}}.
\label{eq:clean-estimate}
\end{equation}
For $t>1$, the reverse update is
\begin{equation}
 Y_{t-1,h}=\mu_h+\gamma_{0,t}(\widehat Y_{0,h}-\mu_h)
 +\gamma_{t,t}(Y_{t,h}-\mu_h)
 +\sqrt{\widetilde\beta_t}A_hz_{t,h},
\label{eq:reverse}
\end{equation}
where $z_{t,h}\sim\mathcal N(0,I_d)$ is independent of the initialization
and other reverse steps. The DDPM posterior coefficients
\cite{ho2020ddpm} are
\begin{equation}
 \gamma_{0,t}=\frac{\beta_t\sqrt{\bar\alpha_{t-1}}}{1-\bar\alpha_t},\quad
 \gamma_{t,t}=\frac{\sqrt{\alpha_t}(1-\bar\alpha_{t-1})}{1-\bar\alpha_t},\quad
 \widetilde\beta_t=\frac{\beta_t(1-\bar\alpha_{t-1})}{1-\bar\alpha_t}.
\label{eq:posterior-coefficients}
\end{equation}
The final step returns $\widehat Y_0$ without additional noise.

With fixed conditional geometry, the change of variables
$A_h^{-1}(Y_{t,h}-\mu_h)$ yields standard Gaussian diffusion, connecting
the construction to conditional-prior and whitening methods
\cite{lee2022priorgrad,yang2026cwgen}. Here the conditional Gaussian is the
reference distribution that initializes a finite-step reverse chain,
whose learned transitions define the final predictive distribution.
With $\bar\alpha_T\approx0.9036$ under the schedule in
Section~\ref{sec:experiments}, the finite-step chain uses an approximate
Gaussian reference initialization. The constraint
$\diag(\Sigma_h)=v_h$ specifies the marginal diffusion-noise variances.

\subsection{Learning and optimization}
\label{sec:training}
\paragraph{Residual estimation}
For minibatch size $B$ and residuals $e_{bh}=Y_{bh}-\mu_{bh}$, the estimator
uses the Gaussian auxiliary score
\begin{equation}
 \mathcal L_{\mathrm{NLL}}=\frac{1}{2BHd}\sum_{b,h}
 [e_{bh}^\top\Sigma_{bh}^{-1}e_{bh}+\log\det\Sigma_{bh}],
\label{eq:nll}
\end{equation}
omitting the additive Gaussian constant.
To distinguish marginal and dependence contributions, write
$\Sigma_h=V_h^{1/2}R_hV_h^{1/2}$ with $V_h=\diag(v_h)$ and
$\widetilde e_h=V_h^{-1/2}e_h$, where $R_h$ is the noise correlation
matrix. The unnormalized per-step score decomposes as
\begin{equation}
 \ell_h=\tfrac12\left[\sum_j\log v_{hj}+\|\widetilde e_h\|_2^2\right]
 +\tfrac12\left[\widetilde e_h^\top(R_h^{-1}-I_d)\widetilde e_h
 +\log\det R_h\right].
\label{eq:nll-decomposition}
\end{equation}
For fixed center and marginals, the second term measures the change from
the diagonal Gaussian score and vanishes at $R_h=I_d$.

\paragraph{Denoising and joint objective}\mbox{}\\
For $t\sim\operatorname{Uniform}\{1,\ldots,T\}$, let
$r_{bh}=\widehat\epsilon_{\theta,bh}-\epsilon_{bh}$ denote the noise
prediction error in Eq.~\eqref{eq:forward}. We balance accuracy before
and after transport:
\begin{equation}
 \mathcal L_{\mathrm{diff}}=\E_{t,\epsilon}\left[
 \frac{1}{2BHd}\sum_{b,h}\|r_{bh}\|_2^2
 +\frac{1}{2BH}\sum_{b,h}
 \frac{\|A_{bh}r_{bh}\|_2^2}{\tr(\Sigma_{bh})}\right].
\label{eq:diff_loss}
\end{equation}
The first term measures white-noise error; the second measures its
data-coordinate effect after covariance transport. Trace normalization
controls the scale of the latter across dimensions and noise magnitudes.
All estimator outputs used in the perturbations, conditioning and error
metric are detached, so this loss updates only the denoiser parameters
$\theta$, including its geometry encoder. The joint objective is
\begin{equation}
 \mathcal L_{\mathrm{joint}}=\mathcal L_{\mathrm{diff}}+
 \mathcal L_{\mathrm{NLL}}+\frac{\|Y-\mu\|_F^2}{BHd}
 +\lambda_c\operatorname{mean}(c^2)
 +\lambda_b\operatorname{mean}(b^2).
\label{eq:joint}
\end{equation}
The residual NLL jointly trains the center, scale and factor estimates;
the center MSE anchors point accuracy, and the last two terms regularize
the correction coefficients. This gradient separation lets the estimator
learn residual statistics while the denoiser learns to reverse the
resulting perturbations.

\paragraph{Staged optimization}
Training proceeds from base estimation to factor warmup and joint
optimization. First, the base predictors minimize
\begin{equation}
 \mathcal L_{\mathrm{base}}=\frac{1}{2BHd}\sum_{b,h,j}
 \left[\log v_{0,bhj}+\frac{(Y_{bhj}-\mu_{0,bhj})^2}{v_{0,bhj}}\right]
 +\frac{\|Y-\mu_0\|_F^2}{BHd}.
\label{eq:base-loss}
\end{equation}
Second, we freeze the validation-selected base, keep $c=b=0$, and train
the factor estimator with $\mathcal L_{\mathrm{NLL}}$ and the denoiser
with $\mathcal L_{\mathrm{diff}}$. Third, we restore the selected warm
checkpoint, enable the zero-initialized correction heads, and optimize
all branches using Eq.~\eqref{eq:joint}.

The base checkpoint is selected by validation $\mathcal L_{\mathrm{base}}$;
warmup and joint checkpoints are selected by validation CRPS with 100
samples per origin. The initial warm checkpoint is included among the
joint-stage candidates. Each stage uses a fresh optimizer; settings and
stage budgets are given in Section~\ref{sec:experiments}.

Algorithm~\ref{alg:training} summarizes the updates. We denote base,
factor, correction, and denoiser parameters by
$\Theta_0$, $\Theta_F$, $\Theta_C$, and $\theta$, respectively.

\begin{algorithm}[tbp]
\caption{FaCoDiff training}
\label{alg:training}
\begin{algorithmic}[1]
\Require Training and validation sets; stage budgets.
\State Initialize parameters; set correction output heads to zero.
\Statex \textbf{Stage 1: Base estimation}
\State Train $\Theta_0$ with $\mathcal L_{\rm base}$; restore its best validation checkpoint.
\Statex \textbf{Stage 2: Factor and denoiser warmup}
\State Freeze $\Theta_0$ and $\Theta_C$; set $c=b=0$.
\For{each factor-warmup minibatch $(X,Y)$}
 \State Compute $(\mu_0,v_0,U,s)$; construct $(D,L)$ via Eq.~\eqref{eq:covariance}.
 \State Use $\sg(\mu_0,v_0,D,L)$ only in the diffusion branch.
 \State Sample $t,\epsilon$; form $Y_t$ via Eq.~\eqref{eq:forward}.
 \State Compute geometry modulations; predict white noise.
 \State Take one optimizer step on $\mathcal L_{\rm NLL}+\mathcal L_{\rm diff}$
 for $(\Theta_F,\theta)$.
\EndFor
\State Restore the warm checkpoint with lowest validation CRPS.
\Statex \textbf{Stage 3: Joint optimization}
\State Unfreeze all branches; enable the zero-initialized correction heads.
\State Include the warm checkpoint as an initial joint-stage candidate.
\For{each joint-optimization minibatch $(X,Y)$}
 \State Compute $(\mu,v,D,L)$ using Eqs.~\eqref{eq:base-predictors}--\eqref{eq:covariance}.
 \State Use $\sg(\mu,v,D,L)$ only in the diffusion branch.
 \State Sample $t,\epsilon$; form $Y_t$ via Eq.~\eqref{eq:forward}.
 \State Compute geometry modulations; predict white noise.
 \State Take one optimizer step on $\mathcal L_{\rm joint}$
 for $(\Theta_0,\Theta_F,\Theta_C,\theta)$.
\EndFor
\State \Return Joint-stage candidate with lowest validation CRPS.
\end{algorithmic}
\end{algorithm}

\subsection{Matrix-free computation and inference}
\label{sec:lowrank}
Both residual estimation and generation operate on $(D_h,L_h)$ without
materializing $\Sigma_h$. Define $K_h=I_r+L_h^\top D_h^{-1}L_h$.
The Woodbury identity and matrix determinant lemma give
\begin{align}
 \Sigma_h^{-1}&=D_h^{-1}
 -D_h^{-1}L_hK_h^{-1}L_h^\top D_h^{-1},\label{eq:woodbury}\\*
 \log\det\Sigma_h&=\sum_j\log\delta_{hj}+\log\det K_h.
\label{eq:logdet}
\end{align}
For noise transport, set $B_h=D_h^{-1/2}L_h$ and diagonalize only
the $r\times r$ Gram matrix:
$B_h^\top B_h=Q_h\Lambda_hQ_h^\top$.
With $G_h=B_hQ_h$, apply
\begin{equation}
\begin{aligned}
 A_hz&=D_h^{1/2}\left[z+G_h\diag(a_{hk})G_h^\top z\right],\\
 a_{hk}&=\frac{1}{\sqrt{1+\lambda_{hk}}+1},
\end{aligned}
\label{eq:transport}
\end{equation}
where $\lambda_{hk}$ are the diagonal entries of $\Lambda_h$.
The bracketed map applies $(I_d+B_hB_h^\top)^{1/2}$ to $z$, so
$A_hA_h^\top=D_h+L_hL_h^\top$; the coefficients remain finite at
zero eigenvalues.

Per forecast step, preparation costs $O(dr^2+r^3)$, storage costs
$O(dr+r^2)$, and each prepared transport costs $O(dr)$ per vector.
In contrast, the dense covariance geometry in CW-Gen \cite{yang2026cwgen}
requires $O(d^2)$ storage, $O(d^3)$ for Cholesky or eigendecomposition,
and $O(d^2)$ per matrix-vector transport. At fixed $r\ll d$, our
covariance storage and operations therefore scale linearly with $d$.
These are covariance-operation costs, separate from the predictor and
denoiser costs; both approaches can prepare their factors once per
history and reuse them across samples. We then run
Eq.~\eqref{eq:reverse} with independent noise draws to obtain $S$ forecasts.

\section{Experiments}
\label{sec:experiments}
\subsection{Datasets and evaluation protocol}
\label{sec:data}
We evaluate \ours{} on nine multivariate time series datasets:
Solar, ECL and Traffic \cite{lai2018lstnet}, Weather and ILI
\cite{wu2021autoformer}, ETTh2 and ETTm2 \cite{zhou2021informer},
and London and Taxi. Table~\ref{tab:data} summarizes their dimensions and
window counts. The history and prediction lengths
are $168\to192$, except for Taxi ($48\to24$) and ILI ($168\to36$).

For London, we use the first 1,000 series in archive order from the
missing-value-free London Smart Meters dataset \cite{godahewa2021monash},
covering 1 May 2012--20 February 2014. Summing consecutive half-hour
readings yields 15,864 hourly observations. Taxi uses all 1,214 series
with 1,488 half-hour observations from the January 2015 training release
of GluonTS taxi\_30min \cite{alexandrov2020gluonts}.

Each series is split chronologically into training, validation and test
sets in a 7:1:2 ratio. We apply per-variable z-score normalization using
training-set statistics and evaluate in the standardized space.
Training windows use stride one; validation and test use non-overlapping
target windows with stride $H$. We retain complete target windows within
each split, allowing historical context from the preceding split.

\begin{table}[tbp]
\centering\tablefont
\setlength{\tabcolsep}{3pt}
\caption{Dataset statistics. $d$ is the number of variables; $N/H$ denotes history/forecast lengths. Train, Val. and Test give window counts.}
\label{tab:data}
\begin{tabular*}{\columnwidth}{@{\extracolsep{\fill}}lccccc@{}}
\toprule
Dataset & $d$ & $N/H$ & Train & Val. & Test \\
\midrule
Solar & 137 & 168/192 & 36433 & 27 & 54 \\
ECL & 321 & 168/192 & 18053 & 13 & 27 \\
Traffic & 862 & 168/192 & 11921 & 9 & 18 \\
London & 1000 & 168/192 & 10745 & 8 & 16 \\
Taxi & 1214 & 48/24 & 970 & 6 & 12 \\
Weather & 21 & 168/192 & 36528 & 27 & 54 \\
ILI & 7 & 168/36 & 473 & 2 & 5 \\
ETTh2 & 7 & 168/192 & 11835 & 9 & 18 \\
ETTm2 & 7 & 168/192 & 48417 & 36 & 72 \\
\bottomrule
\end{tabular*}
\end{table}

All models generate $S=100$ joint forecast samples per test origin and
use a common evaluation pipeline for the six primary metrics below. We report
the mean and sample standard deviation over three independent runs.
Checkpoints are selected using the validation set.

\subsection{Evaluation metrics}
\label{sec:metricdefs}
Let $y_{whj}$ be the standardized observation at test origin $w$, prediction
step $h$, and variable $j$, and let $x^{(s)}_{whj}$ be sample $s$.
There are $W$ origins. Marginal CRPS \cite{gneiting2007scoring} is
\begin{equation}
\begin{aligned}
 \mathrm{CRPS}&=\frac{1}{WHd}\sum_{w,h,j}
 \left[\frac1S\sum_s|x^{(s)}_{whj}-y_{whj}|\right.\\
 &\left.\qquad-\frac{1}{2S^2}\sum_{s,s'}
 |x^{(s)}_{whj}-x^{(s')}_{whj}|\right].
\end{aligned}
\end{equation}
QICE \cite{han2022card} measures deviation from uniform occupancy of ten empirical quantile
intervals. With global count $n_k$ in interval $k$,
\begin{equation}
 \mathrm{QICE}=\frac1{10}\sum_{k=1}^{10}
 \left|\frac{n_k}{WHd}-0.1\right|.
\end{equation}
We use linear percentile interpolation and strict quantile exceedances;
observations outside the sample range enter the adjacent edge intervals.
Counts are pooled across all origins before computing deviations.

Energy \cite{gneiting2007scoring} scores the joint $d$-variable forecast
at each prediction step:
\begin{equation}
 \mathrm{Energy}=\frac1{WH}\sum_{w,h}
 \left[\frac1S\sum_s\|x^{(s)}_{wh}-y_{wh}\|_2
 -\frac{1}{2S^2}\sum_{s,s'}
 \|x^{(s)}_{wh}-x^{(s')}_{wh}\|_2\right].
\end{equation}
Variogram \cite{scheuerer2015variogram} uses exponent $p=1/2$ and equal normalized weights over all
unordered variable pairs:
\begin{equation}
 \mathrm{VS}_{p}=\frac1{WH\binom d2}\sum_{w,h}\sum_{i<j}
 \left[|y_{whi}-y_{whj}|^p-
 \frac1S\sum_s|x^{(s)}_{whi}-x^{(s)}_{whj}|^p\right]^2.
\end{equation}
Energy uses the unnormalized Euclidean norm; both joint scores evaluate
the $d$-variable forecast separately at each prediction step. Sample-pair averages use
$S^2$, including zero-distance diagonal pairs. Finally, with sample mean $\bar x_{whj}=S^{-1}\sum_sx^{(s)}_{whj}$,
\begin{equation}
\begin{aligned}
 \mathrm{MAE}&=\frac1{WHd}\sum_{w,h,j}|\bar x_{whj}-y_{whj}|,\\
 \mathrm{MSE}&=\frac1{WHd}\sum_{w,h,j}(\bar x_{whj}-y_{whj})^2.
\end{aligned}
\end{equation}
Thus point errors assess the generated sample mean.
All six metrics are minimized.
For the Solar and Weather component and pairing analyses, we additionally
evaluate a standardized linear aggregate. At each prediction step, define
$a_{wh}=d^{-1/2}\sum_j y_{whj}$ and
$a^{(s)}_{wh}=d^{-1/2}\sum_j x^{(s)}_{whj}$.
Let $m_a$ and $s_a$ be the mean and population standard deviation of the
corresponding aggregate over the training prefix only, and set
\begin{equation}
 \widetilde a_{wh}=\frac{a_{wh}-m_a}{s_a},\qquad
 \widetilde a^{(s)}_{wh}=\frac{a^{(s)}_{wh}-m_a}{s_a}.
 \label{eq:aggregate-score}
\end{equation}
Aggregate CRPS applies the same empirical CRPS formula to these scalar
samples and observations, averaging over origins and prediction steps.
All variants use the same training-fitted aggregate scaler. Lower aggregate
CRPS is better.

\subsection{Implementation details}
The main forecasting benchmarks follow Section~\ref{sec:training}.
Base pretraining uses at most 20 epochs with patience five. Factor warmup
and joint optimization use at most 10 and 20 epochs, respectively, with
run-specific budgets matched between Full and its controls.
Adam uses learning rate $10^{-4}$ for every parameter group, with global
gradient clipping at norm one and fresh optimizer state at stage boundaries.
Training batch size is 32. Validation and test evaluate one origin at a time,
generating samples in chunks of ten. The two correction penalties have
weight $10^{-3}$ each. Both temporal factor and correction encoders use
two-layer MLPs of width 64 with SiLU activations and layer normalization.
The denoiser has three hidden layers of width 128. Its geometry encoder
has a SiLU hidden layer of width 128 and outputs the six modulation
vectors in Eq.~\eqref{eq:geometry-film}. The white-coordinate and
trace-normalized transported-error terms receive equal weight.

We use eight factor columns ($r=8$) on all main-table datasets.
Diffusion training and ancestral sampling both use $T=20$ steps, as in
NsDiff \cite{ye2025nsdiff}. We use a linear schedule from
$\beta_1=10^{-4}$ to $\beta_T=0.01$.
The conditional Gaussian in Section~\ref{sec:diffusion} initializes this
short reverse chain. Marginal variance is
floored at $10^{-6}$, and the small linear systems use numerical diagonal
stabilization. Validation sampling uses a fixed random seed for each run.

\paragraph{Baselines}
We compare with DeepVAR \cite{salinas2019copula}, TMDM \cite{li2024tmdm}, NsDiff \cite{ye2025nsdiff},
CW-NsDiff \cite{yang2026cwgen}, and
DiffPTS \cite{ye2026diffpts}. All baselines use public implementations
with model-specific training settings and the common evaluation protocol.
DeepVAR uses the PyTorchTS low-rank Gaussian implementation with rank five,
100 training epochs and 50 updates per epoch.
CW-NsDiff is the NsDiff instantiation of CW-Gen. It uses the official
high-dimensional accelerated variant on Solar and ECL
and the full JMCE loss on the remaining evaluated datasets.
Entries unavailable within the memory and runtime budget are marked --.

\subsection{Main forecasting results}
\label{sec:main-results}
\paragraph{Joint forecast quality}
The clearest cross-dataset advantage is in dependence-sensitive forecast
quality: \ours{} achieves the lowest mean Variogram score on eight of
nine datasets in Table~\ref{tab:probability}. On Traffic, the reduction
relative to DiffPTS~\cite{ye2026diffpts} is 18.88\%.

\paragraph{Marginal and point forecasting}
This joint-score advantage is accompanied by competitive marginal and
point forecasts. Across all six metrics, \ours{} ranks among the top two
on 43 of the 54 dataset--metric combinations. For example, CRPS is
40.99\% lower than NsDiff on Solar and 11.27\% lower than DiffPTS on
Traffic. The gains therefore extend beyond dependence-sensitive scoring
to the quality of individual-variable predictions.

\paragraph{Variation across datasets}
The advantage varies with the dataset and evaluation criterion. TMDM
achieves the lowest means on all six metrics on Weather. On London and
Taxi, \ours{} has the lowest CRPS, QICE and Variogram, while NsDiff has
the lowest Energy. These differences distinguish the broad Variogram
advantage from improvements in other aspects of forecast quality.

\begin{table*}[p]
\centering
\rmfamily\fontsize{7.2}{8.6}\selectfont
\setlength{\tabcolsep}{3pt}
\renewcommand{\arraystretch}{0.96}
\caption{Forecasting results on nine datasets. Results are mean $\pm$ standard deviation over three runs. Lower is better; red and blue mark the best and second-best available means. -- denotes unavailable results.}
\label{tab:probability}
\begin{tabular*}{\textwidth}{@{\extracolsep{\fill}}llcccccc@{}}
\toprule
Dataset & Model & CRPS & QICE & Energy & Variogram & MAE & MSE \\
\midrule
Solar & DeepVAR & $0.5039\pm 0.0013$ & $0.0788\pm 0.0001$ & $6.2431\pm 0.0090$ & $\secondmean{0.1606}\pm 0.0059$ & $0.6490\pm 0.0034$ & $0.7658\pm 0.0026$ \\
 & TMDM & $0.3567\pm 0.0032$ & $0.0901\pm 0.0007$ & $5.2124\pm 0.0386$ & $0.6747\pm 0.0039$ & $\secondmean{0.3266}\pm 0.0047$ & $\secondmean{0.2793}\pm 0.0078$ \\
 & NsDiff & $0.3136\pm 0.0141$ & $0.0687\pm 0.0016$ & $4.4388\pm 0.1722$ & $0.4645\pm 0.0073$ & $0.3329\pm 0.0189$ & $0.2850\pm 0.0280$ \\
 & CW-NsDiff & $\secondmean{0.3116}\pm 0.0089$ & $0.0645\pm 0.0049$ & $\secondmean{4.1631}\pm 0.0830$ & $0.2381\pm 0.0034$ & $0.3630\pm 0.0211$ & $0.3254\pm 0.0096$ \\
 & DiffPTS & $0.3453\pm 0.0329$ & $\firstmean{0.0357}\pm 0.0173$ & $4.8937\pm 0.4662$ & $0.3351\pm 0.0343$ & $0.4958\pm 0.0563$ & $0.8730\pm 0.1842$ \\
 & FaCoDiff & $\firstmean{0.1851}\pm 0.0129$ & $\secondmean{0.0587}\pm 0.0190$ & $\firstmean{2.4731}\pm 0.1569$ & $\firstmean{0.0557}\pm 0.0016$ & $\firstmean{0.2574}\pm 0.0118$ & $\firstmean{0.2309}\pm 0.0194$ \\
\midrule
ECL & DeepVAR & $0.5796\pm 0.0040$ & $\secondmean{0.0187}\pm 0.0033$ & $12.7429\pm 0.1092$ & $0.1397\pm 0.0087$ & $0.8263\pm 0.0050$ & $1.0506\pm 0.0134$ \\
 & TMDM & $0.4503\pm 0.0032$ & $0.1056\pm 0.0012$ & $10.3029\pm 0.0637$ & $0.3703\pm 0.0004$ & $0.3489\pm 0.0099$ & $0.2425\pm 0.0109$ \\
 & NsDiff & $0.2903\pm 0.0016$ & $0.0797\pm 0.0007$ & $6.6882\pm 0.0288$ & $0.1513\pm 0.0024$ & $0.2964\pm 0.0009$ & $0.1939\pm 0.0023$ \\
 & CW-NsDiff & $0.3413\pm 0.0148$ & $0.0600\pm 0.0064$ & $7.5079\pm 0.3164$ & $0.1088\pm 0.0042$ & $0.4046\pm 0.0316$ & $0.3011\pm 0.0425$ \\
 & DiffPTS & $\secondmean{0.2142}\pm 0.0016$ & $\firstmean{0.0173}\pm 0.0024$ & $\firstmean{5.3193}\pm 0.0492$ & $\secondmean{0.0706}\pm 0.0002$ & $\secondmean{0.2908}\pm 0.0038$ & $\secondmean{0.1881}\pm 0.0043$ \\
 & FaCoDiff & $\firstmean{0.2124}\pm 0.0031$ & $0.0206\pm 0.0047$ & $\secondmean{5.3422}\pm 0.0763$ & $\firstmean{0.0687}\pm 0.0007$ & $\firstmean{0.2847}\pm 0.0046$ & $\firstmean{0.1854}\pm 0.0065$ \\
\midrule
Traffic & DeepVAR & $0.6142\pm 0.0023$ & $0.0385\pm 0.0017$ & $24.2840\pm 0.1782$ & $0.2462\pm 0.0159$ & $0.7996\pm 0.0010$ & $1.5187\pm 0.0026$ \\
 & TMDM & $0.5630\pm 0.0029$ & $0.1075\pm 0.0011$ & $22.0758\pm 0.0759$ & $0.6701\pm 0.0008$ & $0.4325\pm 0.0113$ & $0.7150\pm 0.0105$ \\
 & NsDiff & $0.3764\pm 0.0058$ & $0.0851\pm 0.0026$ & $16.8018\pm 0.1950$ & $0.2943\pm 0.0042$ & $0.3747\pm 0.0158$ & $0.6511\pm 0.0252$ \\
 & CW-NsDiff & -- & -- & -- & -- & -- & -- \\
 & DiffPTS & $\secondmean{0.2776}\pm 0.0012$ & $\secondmean{0.0360}\pm 0.0052$ & $\secondmean{14.1006}\pm 0.1004$ & $\secondmean{0.1322}\pm 0.0036$ & $\secondmean{0.3501}\pm 0.0031$ & $\secondmean{0.5985}\pm 0.0083$ \\
 & FaCoDiff & $\firstmean{0.2463}\pm 0.0025$ & $\firstmean{0.0188}\pm 0.0008$ & $\firstmean{14.0913}\pm 0.2352$ & $\firstmean{0.1072}\pm 0.0011$ & $\firstmean{0.3168}\pm 0.0032$ & $\firstmean{0.5691}\pm 0.0097$ \\
\midrule
London & DeepVAR & $0.5991\pm 0.0265$ & $0.0442\pm 0.0052$ & $28.6807\pm 1.0316$ & $0.2766\pm 0.0261$ & $0.7554\pm 0.0199$ & $1.6715\pm 0.1017$ \\
 & TMDM & $0.6558\pm 0.0022$ & $0.0821\pm 0.0008$ & $27.0359\pm 0.0809$ & $0.5519\pm 0.0004$ & $0.6732\pm 0.0090$ & $1.1326\pm 0.0135$ \\
 & NsDiff & $0.5038\pm 0.0077$ & $0.0486\pm 0.0034$ & $\firstmean{22.6569}\pm 0.3350$ & $0.2791\pm 0.0014$ & $\secondmean{0.6229}\pm 0.0133$ & $\firstmean{1.0454}\pm 0.0371$ \\
 & CW-NsDiff & -- & -- & -- & -- & -- & -- \\
 & DiffPTS & $\secondmean{0.4944}\pm 0.0011$ & $\secondmean{0.0427}\pm 0.0042$ & $\secondmean{22.7260}\pm 0.0512$ & $\secondmean{0.2553}\pm 0.0042$ & $0.6371\pm 0.0023$ & $\secondmean{1.0858}\pm 0.0067$ \\
 & FaCoDiff & $\firstmean{0.4648}\pm 0.0033$ & $\firstmean{0.0233}\pm 0.0008$ & $22.7903\pm 0.1838$ & $\firstmean{0.2135}\pm 0.0018$ & $\firstmean{0.6166}\pm 0.0056$ & $1.0883\pm 0.0191$ \\
\midrule
Taxi & DeepVAR & $0.5724\pm 0.0019$ & $\secondmean{0.0184}\pm 0.0013$ & $24.0245\pm 0.1304$ & $\secondmean{0.1787}\pm 0.0023$ & $0.7987\pm 0.0055$ & $1.0242\pm 0.0153$ \\
 & TMDM & $0.6080\pm 0.0043$ & $0.0884\pm 0.0016$ & $26.5789\pm 0.1810$ & $0.6324\pm 0.0019$ & $0.5867\pm 0.0143$ & $0.6267\pm 0.0240$ \\
 & NsDiff & $0.4298\pm 0.0024$ & $0.0218\pm 0.0009$ & $\firstmean{18.9038}\pm 0.1228$ & $0.1807\pm 0.0012$ & $\firstmean{0.5666}\pm 0.0028$ & $\firstmean{0.6032}\pm 0.0025$ \\
 & CW-NsDiff & -- & -- & -- & -- & -- & -- \\
 & DiffPTS & $\secondmean{0.4278}\pm 0.0028$ & $0.0296\pm 0.0035$ & $18.9582\pm 0.0820$ & $0.1835\pm 0.0046$ & $\secondmean{0.5688}\pm 0.0090$ & $\secondmean{0.6077}\pm 0.0097$ \\
 & FaCoDiff & $\firstmean{0.4246}\pm 0.0173$ & $\firstmean{0.0152}\pm 0.0013$ & $\secondmean{18.9423}\pm 0.6943$ & $\firstmean{0.1396}\pm 0.0051$ & $0.5828\pm 0.0235$ & $0.6438\pm 0.0529$ \\
\midrule
Weather & DeepVAR & $0.3539\pm 0.0852$ & $0.0333\pm 0.0051$ & $2.3485\pm 0.3778$ & $0.1627\pm 0.0886$ & $0.4640\pm 0.0792$ & $0.5606\pm 0.0897$ \\
 & TMDM & $\firstmean{0.2095}\pm 0.0085$ & $\firstmean{0.0280}\pm 0.0067$ & $\firstmean{1.4291}\pm 0.0759$ & $\firstmean{0.0792}\pm 0.0047$ & $\firstmean{0.2748}\pm 0.0097$ & $\firstmean{0.2333}\pm 0.0181$ \\
 & NsDiff & $0.2409\pm 0.0029$ & $0.0335\pm 0.0017$ & $1.5013\pm 0.0158$ & $0.0943\pm 0.0004$ & $0.2964\pm 0.0045$ & $0.2493\pm 0.0062$ \\
 & CW-NsDiff & $0.3203\pm 0.0141$ & $0.0345\pm 0.0068$ & $1.8729\pm 0.0824$ & $0.1363\pm 0.0093$ & $0.4091\pm 0.0238$ & $0.3665\pm 0.0378$ \\
 & DiffPTS & $0.2154\pm 0.0049$ & $0.0354\pm 0.0067$ & $\secondmean{1.4423}\pm 0.0341$ & $\secondmean{0.0830}\pm 0.0014$ & $0.2865\pm 0.0076$ & $\secondmean{0.2378}\pm 0.0131$ \\
 & FaCoDiff & $\secondmean{0.2151}\pm 0.0048$ & $\secondmean{0.0311}\pm 0.0068$ & $1.4873\pm 0.0239$ & $0.0836\pm 0.0019$ & $\secondmean{0.2815}\pm 0.0038$ & $0.2482\pm 0.0050$ \\
\midrule
ILI & DeepVAR & $1.4226\pm 0.0793$ & $0.1169\pm 0.0220$ & $4.3472\pm 0.2563$ & $0.3741\pm 0.0630$ & $1.5848\pm 0.0902$ & $5.6123\pm 0.7076$ \\
 & TMDM & $0.9295\pm 0.1540$ & $\secondmean{0.0387}\pm 0.0063$ & $2.8730\pm 0.4298$ & $0.2597\pm 0.0017$ & $1.1851\pm 0.1894$ & $3.2823\pm 0.9993$ \\
 & NsDiff & $0.8688\pm 0.0440$ & $0.0535\pm 0.0090$ & $2.6917\pm 0.1172$ & $0.3043\pm 0.0055$ & $1.0712\pm 0.0479$ & $2.7492\pm 0.2598$ \\
 & CW-NsDiff & $\firstmean{0.6937}\pm 0.0271$ & $\firstmean{0.0293}\pm 0.0080$ & $\firstmean{2.1920}\pm 0.1035$ & $\secondmean{0.2079}\pm 0.0218$ & $\secondmean{0.9165}\pm 0.0275$ & $\secondmean{1.8601}\pm 0.1272$ \\
 & DiffPTS & $0.8556\pm 0.0484$ & $0.0732\pm 0.0109$ & $2.6337\pm 0.1364$ & $0.2101\pm 0.0182$ & $1.0031\pm 0.0701$ & $2.7173\pm 0.3782$ \\
 & FaCoDiff & $\secondmean{0.7097}\pm 0.0759$ & $0.0678\pm 0.0149$ & $\secondmean{2.2280}\pm 0.2118$ & $\firstmean{0.1850}\pm 0.0131$ & $\firstmean{0.8570}\pm 0.0797$ & $\firstmean{1.8307}\pm 0.3524$ \\
\midrule
ETTh2 & DeepVAR & $0.3549\pm 0.0225$ & $0.0389\pm 0.0058$ & $1.2718\pm 0.0950$ & $0.1276\pm 0.0061$ & $0.4639\pm 0.0294$ & $0.5621\pm 0.0406$ \\
 & TMDM & $0.2846\pm 0.0240$ & $\secondmean{0.0191}\pm 0.0020$ & $0.9040\pm 0.0754$ & $0.0847\pm 0.0089$ & $0.3838\pm 0.0296$ & $0.2840\pm 0.0572$ \\
 & NsDiff & $0.2712\pm 0.0113$ & $\firstmean{0.0131}\pm 0.0034$ & $0.8899\pm 0.0358$ & $0.0789\pm 0.0059$ & $0.3678\pm 0.0126$ & $0.2839\pm 0.0305$ \\
 & CW-NsDiff & $0.2748\pm 0.0061$ & $0.0205\pm 0.0012$ & $0.8862\pm 0.0162$ & $0.0786\pm 0.0024$ & $0.3652\pm 0.0070$ & $0.2819\pm 0.0181$ \\
 & DiffPTS & $\firstmean{0.2629}\pm 0.0169$ & $0.0350\pm 0.0065$ & $\firstmean{0.8467}\pm 0.0477$ & $\secondmean{0.0775}\pm 0.0090$ & $\firstmean{0.3435}\pm 0.0158$ & $\firstmean{0.2401}\pm 0.0167$ \\
 & FaCoDiff & $\secondmean{0.2656}\pm 0.0193$ & $0.0218\pm 0.0097$ & $\secondmean{0.8618}\pm 0.0624$ & $\firstmean{0.0771}\pm 0.0069$ & $\secondmean{0.3555}\pm 0.0253$ & $\secondmean{0.2656}\pm 0.0419$ \\
\midrule
ETTm2 & DeepVAR & $0.3348\pm 0.0321$ & $0.0387\pm 0.0163$ & $1.1778\pm 0.1444$ & $0.1159\pm 0.0042$ & $0.4507\pm 0.0173$ & $0.4881\pm 0.0348$ \\
 & TMDM & $\secondmean{0.2112}\pm 0.0031$ & $\secondmean{0.0187}\pm 0.0012$ & $\secondmean{0.6867}\pm 0.0124$ & $\secondmean{0.0572}\pm 0.0020$ & $\secondmean{0.2795}\pm 0.0030$ & $\secondmean{0.1726}\pm 0.0078$ \\
 & NsDiff & $0.2228\pm 0.0068$ & $\firstmean{0.0182}\pm 0.0015$ & $0.7196\pm 0.0209$ & $0.0600\pm 0.0020$ & $0.2950\pm 0.0086$ & $0.1865\pm 0.0129$ \\
 & CW-NsDiff & $0.2226\pm 0.0048$ & $0.0226\pm 0.0023$ & $0.7166\pm 0.0169$ & $0.0597\pm 0.0018$ & $0.2943\pm 0.0066$ & $0.1808\pm 0.0077$ \\
 & DiffPTS & $0.2658\pm 0.0440$ & $0.0418\pm 0.0157$ & $0.8652\pm 0.1364$ & $0.0852\pm 0.0224$ & $0.3386\pm 0.0368$ & $0.2210\pm 0.0436$ \\
 & FaCoDiff & $\firstmean{0.2081}\pm 0.0039$ & $0.0188\pm 0.0015$ & $\firstmean{0.6756}\pm 0.0103$ & $\firstmean{0.0551}\pm 0.0013$ & $\firstmean{0.2748}\pm 0.0048$ & $\firstmean{0.1681}\pm 0.0066$ \\
\bottomrule
\end{tabular*}
\end{table*}

\subsection{Forecast visualization}
\label{sec:forecast-case}
\begin{figure*}[pos=tbp]
\centering
\includegraphics[width=\textwidth]{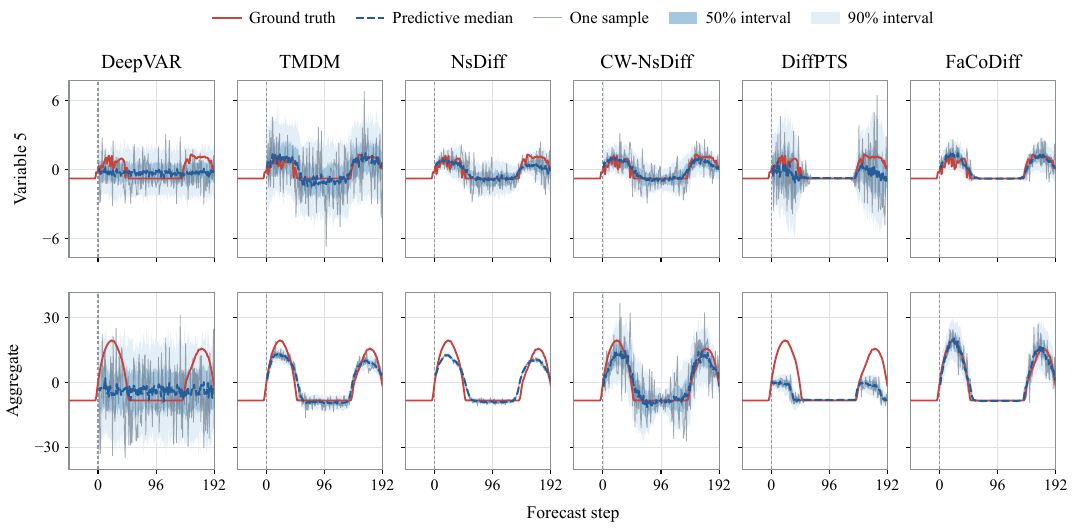}
\caption{A Solar forecasting example. Top: a single variable; bottom: the normalized aggregate $137^{-1/2}\sum_{j=1}^{137}Y_j$ of training-standardized variables. Red: observations; dashed blue: predictive median; gray: one generated trajectory. Shading indicates pointwise 50\% and 90\% prediction intervals.}
\label{fig:real-case}
\end{figure*}

Figure~\ref{fig:real-case} compares FaCoDiff with DeepVAR and the diffusion
baselines on a Solar forecast with irregular intraday fluctuations.
The upper row compares individual-variable trajectories;
the lower row aggregates all variables within each joint sample.
Predictive medians, one sampled trajectory and central prediction
intervals reveal both the forecast profile and its uncertainty. In this
example, FaCoDiff tracks the two aggregate peaks with less attenuation
than NsDiff and DiffPTS, while its intervals contract during low-output
periods.

\subsection{Ablation study}
\label{sec:controls}
We separate the contributions of factor covariance, marginal corrections
and diffusion generation on Solar and Weather. Table~\ref{tab:ablation-components}
crosses covariance on/off with four correction settings: none, Location,
Scale and both. Within each seed,
all variants start from the same pretrained base checkpoint. Covariance controls
share initialization and random seeds, with factor-warmup and joint-training
budgets matched to Full. Location/Scale controls
change only the enabled correction heads within this common training recipe.
Each variant selects its checkpoint by validation CRPS, including the
joint-stage initial checkpoint. Full is the main-benchmark model reported in Table~\ref{tab:probability}.

\textit{Diagonal} removes factor covariance and both corrections while
retaining residual NLL, the denoising objective, gradient routing, and matched
training budget. \textit{Diagonal+L}, \textit{Diagonal+S}, and
\textit{Diagonal+L+S} retain the conditional correction basis with Location,
Scale, or both enabled, respectively, but use $\diag(v)$ as noise covariance.
\textit{Factor-only} retains conditional covariance without either correction;
\textit{Factor+L} and \textit{Factor+S} enable Location or Scale separately.
\textit{Full} enables all three outputs. Disabled heads receive no
branch-specific losses or updates. All diagonal variants retain FaCoDiff's
residual-learning recipe, distinguishing these component controls from
the native NsDiff benchmark.

\begin{table*}[tbp]
\centering
\rmfamily\fontsize{8}{9.6}\selectfont
\setlength{\tabcolsep}{3pt}
\renewcommand{\arraystretch}{0.96}
\caption{Component ablations on Solar and Weather. Each cell gives the mean above its standard deviation over three runs. Agg. CRPS scores the standardized linear aggregate in Eq.~\eqref{eq:aggregate-score}. D: Diagonal; F: Factor-only; L: Location; S: Scale. Full includes F, L and S. Lower is better; red marks the best mean.}
\label{tab:ablation-components}
\begin{tabular*}{\textwidth}{@{\extracolsep{\fill}}llccccccc@{}}
\toprule
Dataset & Variant & CRPS & QICE & Energy & Variogram & MAE & MSE & Agg. CRPS \\
\midrule
Solar & D & \shortstack[c]{$0.19818$\\$\pm 0.00436$} & \shortstack[c]{$0.06004$\\$\pm 0.01745$} & \shortstack[c]{$2.70474$\\$\pm 0.04872$} & \shortstack[c]{$0.06981$\\$\pm 0.00145$} & \shortstack[c]{$0.27293$\\$\pm 0.00524$} & \shortstack[c]{$0.25247$\\$\pm 0.00447$} & \shortstack[c]{$0.22245$\\$\pm 0.00511$} \\
 & D+L & \shortstack[c]{$0.19427$\\$\pm 0.00692$} & \shortstack[c]{$0.06009$\\$\pm 0.00746$} & \shortstack[c]{$2.64126$\\$\pm 0.07711$} & \shortstack[c]{$0.06432$\\$\pm 0.00132$} & \shortstack[c]{$0.26827$\\$\pm 0.00790$} & \shortstack[c]{$0.24879$\\$\pm 0.01040$} & \shortstack[c]{$0.22031$\\$\pm 0.00929$} \\
 & D+S & \shortstack[c]{$0.20005$\\$\pm 0.00578$} & \shortstack[c]{$0.06772$\\$\pm 0.02341$} & \shortstack[c]{$2.72537$\\$\pm 0.06504$} & \shortstack[c]{$0.06914$\\$\pm 0.00216$} & \shortstack[c]{$0.27576$\\$\pm 0.00692$} & \shortstack[c]{$0.25319$\\$\pm 0.00722$} & \shortstack[c]{$0.22502$\\$\pm 0.00780$} \\
 & D+L+S & \shortstack[c]{$0.19065$\\$\pm 0.00770$} & \shortstack[c]{$0.06190$\\$\pm 0.00931$} & \shortstack[c]{$2.59665$\\$\pm 0.09022$} & \shortstack[c]{$0.06423$\\$\pm 0.00355$} & \shortstack[c]{$0.26465$\\$\pm 0.00951$} & \shortstack[c]{$0.23911$\\$\pm 0.00918$} & \shortstack[c]{$0.21723$\\$\pm 0.01023$} \\
 & F & \shortstack[c]{$0.19272$\\$\pm 0.00172$} & \shortstack[c]{$\best{0.04439}$\\$\pm 0.00559$} & \shortstack[c]{$2.61648$\\$\pm 0.01726$} & \shortstack[c]{$0.06889$\\$\pm 0.00700$} & \shortstack[c]{$0.26835$\\$\pm 0.00331$} & \shortstack[c]{$0.25058$\\$\pm 0.00128$} & \shortstack[c]{$0.17764$\\$\pm 0.01460$} \\
 & F+L & \shortstack[c]{$0.18783$\\$\pm 0.00733$} & \shortstack[c]{$0.05028$\\$\pm 0.03005$} & \shortstack[c]{$2.50693$\\$\pm 0.08645$} & \shortstack[c]{$0.06117$\\$\pm 0.00967$} & \shortstack[c]{$0.26429$\\$\pm 0.00590$} & \shortstack[c]{$0.23209$\\$\pm 0.00582$} & \shortstack[c]{$0.17177$\\$\pm 0.01264$} \\
 & F+S & \shortstack[c]{$0.19213$\\$\pm 0.00254$} & \shortstack[c]{$0.04715$\\$\pm 0.00297$} & \shortstack[c]{$2.60817$\\$\pm 0.02247$} & \shortstack[c]{$0.06830$\\$\pm 0.00601$} & \shortstack[c]{$0.26802$\\$\pm 0.00274$} & \shortstack[c]{$0.24766$\\$\pm 0.00626$} & \shortstack[c]{$0.17680$\\$\pm 0.01605$} \\
 & Full & \shortstack[c]{$\best{0.18506}$\\$\pm 0.01290$} & \shortstack[c]{$0.05868$\\$\pm 0.01903$} & \shortstack[c]{$\best{2.47311}$\\$\pm 0.15687$} & \shortstack[c]{$\best{0.05575}$\\$\pm 0.00157$} & \shortstack[c]{$\best{0.25740}$\\$\pm 0.01184$} & \shortstack[c]{$\best{0.23088}$\\$\pm 0.01935$} & \shortstack[c]{$\best{0.16908}$\\$\pm 0.01938$} \\
\midrule
Weather & D & \shortstack[c]{$0.22212$\\$\pm 0.00865$} & \shortstack[c]{$0.03779$\\$\pm 0.01321$} & \shortstack[c]{$1.51724$\\$\pm 0.04848$} & \shortstack[c]{$0.08634$\\$\pm 0.00232$} & \shortstack[c]{$0.28398$\\$\pm 0.00730$} & \shortstack[c]{$0.24954$\\$\pm 0.00829$} & \shortstack[c]{$0.20887$\\$\pm 0.00488$} \\
 & D+L & \shortstack[c]{$0.22251$\\$\pm 0.01413$} & \shortstack[c]{$0.03967$\\$\pm 0.01770$} & \shortstack[c]{$1.51923$\\$\pm 0.07893$} & \shortstack[c]{$0.08629$\\$\pm 0.00416$} & \shortstack[c]{$0.28281$\\$\pm 0.01060$} & \shortstack[c]{$0.24663$\\$\pm 0.01286$} & \shortstack[c]{$0.20778$\\$\pm 0.01077$} \\
 & D+S & \shortstack[c]{$0.22117$\\$\pm 0.00669$} & \shortstack[c]{$0.03822$\\$\pm 0.01406$} & \shortstack[c]{$1.51038$\\$\pm 0.03604$} & \shortstack[c]{$0.08572$\\$\pm 0.00175$} & \shortstack[c]{$0.28274$\\$\pm 0.00646$} & \shortstack[c]{$0.24667$\\$\pm 0.00603$} & \shortstack[c]{$0.20872$\\$\pm 0.00464$} \\
 & D+L+S & \shortstack[c]{$0.22455$\\$\pm 0.01454$} & \shortstack[c]{$0.04066$\\$\pm 0.01860$} & \shortstack[c]{$1.53097$\\$\pm 0.08361$} & \shortstack[c]{$0.08670$\\$\pm 0.00405$} & \shortstack[c]{$0.28614$\\$\pm 0.00981$} & \shortstack[c]{$0.25025$\\$\pm 0.01380$} & \shortstack[c]{$0.21168$\\$\pm 0.01027$} \\
 & F & \shortstack[c]{$0.21997$\\$\pm 0.00806$} & \shortstack[c]{$0.03586$\\$\pm 0.00653$} & \shortstack[c]{$1.51463$\\$\pm 0.04141$} & \shortstack[c]{$0.08526$\\$\pm 0.00310$} & \shortstack[c]{$0.28391$\\$\pm 0.00591$} & \shortstack[c]{$0.25294$\\$\pm 0.00829$} & \shortstack[c]{$0.20401$\\$\pm 0.00324$} \\
 & F+L & \shortstack[c]{$0.21851$\\$\pm 0.01007$} & \shortstack[c]{$0.03330$\\$\pm 0.01278$} & \shortstack[c]{$1.49914$\\$\pm 0.05043$} & \shortstack[c]{$0.08473$\\$\pm 0.00357$} & \shortstack[c]{$0.28297$\\$\pm 0.00402$} & \shortstack[c]{$0.25008$\\$\pm 0.00654$} & \shortstack[c]{$0.20077$\\$\pm 0.01084$} \\
 & F+S & \shortstack[c]{$0.21835$\\$\pm 0.00710$} & \shortstack[c]{$0.03757$\\$\pm 0.01101$} & \shortstack[c]{$1.49580$\\$\pm 0.03482$} & \shortstack[c]{$0.08394$\\$\pm 0.00243$} & \shortstack[c]{$\best{0.28008}$\\$\pm 0.00232$} & \shortstack[c]{$\best{0.24566}$\\$\pm 0.00480$} & \shortstack[c]{$0.20186$\\$\pm 0.00417$} \\
 & Full & \shortstack[c]{$\best{0.21513}$\\$\pm 0.00480$} & \shortstack[c]{$\best{0.03108}$\\$\pm 0.00685$} & \shortstack[c]{$\best{1.48730}$\\$\pm 0.02395$} & \shortstack[c]{$\best{0.08364}$\\$\pm 0.00190$} & \shortstack[c]{$0.28153$\\$\pm 0.00381$} & \shortstack[c]{$0.24823$\\$\pm 0.00495$} & \shortstack[c]{$\best{0.19700}$\\$\pm 0.00349$} \\
\bottomrule
\end{tabular*}
\end{table*}

\paragraph{Effect of factor covariance}
Factor covariance improves both joint-score means under all four
correction settings on both datasets (Table~\ref{tab:ablation-components}).
Without corrections, F lowers Energy and Variogram by 3.26\% and 1.32\%
on Solar, and by 0.17\% and 1.25\% on Weather, relative to D.
With both corrections enabled, Full lowers these scores by 4.76\% and
13.20\% on Solar and by 2.85\% and 3.53\% on Weather, relative to D+L+S.
Full also improves all six mean scores over D+L+S on each dataset.
Relative
changes in these component comparisons use ratios of three-run mean scores.
The aggregate scores extend this comparison to a cross-variable linear
target. Factor covariance lowers mean aggregate CRPS under all four
correction settings: by 20.14--22.17\% on Solar and 2.33--6.93\% on Weather.
All 12 seed-matched factor/diagonal comparisons improve on Solar, as do
11 of 12 on Weather. Full has the lowest aggregate-CRPS mean among the
eight variants on both datasets, reducing it by 22.17\% and 6.93\% relative
to D+L+S, respectively. These results show that the covariance benefit
extends to aggregate probabilistic prediction.

\paragraph{Effect of marginal corrections}
Location and Scale strengthen the factor model in complementary ways.
On Solar, adding Location to F lowers MSE by 7.38\% and Variogram by
11.20\%; enabling Scale on top of F+L further lowers Energy and
Variogram by 1.35\% and 8.86\%. Full has the best means on five of the six primary metrics,
while F has the lowest QICE. On Weather, Full has the lowest CRPS, QICE,
Energy and Variogram, and F+S has the lowest point errors. The complete
configuration therefore gives the strongest joint-score means on both
datasets, with the individual controls identifying the contributions of
center adjustment and scale modulation.

\paragraph{Effect of geometry-conditioned denoising}
Table~\ref{tab:denoiser-ablation} crosses explicit geometry conditioning
(G) with the mixed-error objective (T) in Eq.~\eqref{eq:diff_loss}.
Without T, training uses white-noise MSE alone. All four variants retain
factor covariance and both corrections, using the same pretrained
base checkpoint and matched training budgets within each seed; G+T is Full. With the mixed objective fixed, geometry conditioning
reduces CRPS, Energy and Variogram by 3.23\%, 2.97\% and 6.68\% on Solar,
and by 2.55\%, 1.84\% and 1.93\% on Weather.
The combined configuration has the lowest CRPS, QICE and Energy means
among the four variants on both datasets. Variogram favors the
white-error control on Solar and the combined configuration on Weather.
These results support the joint use of geometry conditioning and the
transport-aware objective.

\begin{table*}[tbp]
\centering
\rmfamily\fontsize{7.2}{8.6}\selectfont
\setlength{\tabcolsep}{3pt}
\renewcommand{\arraystretch}{0.96}
\caption{Denoising ablations on Solar and Weather. G: geometry conditioning; T: mixed denoising objective in Eq.~\eqref{eq:diff_loss}. Results are mean $\pm$ standard deviation over three runs. Lower is better; red marks the lowest mean.}
\label{tab:denoiser-ablation}
\begin{tabular*}{\textwidth}{@{\extracolsep{\fill}}lcccccccc@{}}
\toprule
Dataset & G & T & CRPS & QICE & Energy & Variogram & MAE & MSE \\
\midrule
Solar & -- & -- & $0.19083\pm 0.00670$ & $0.07180\pm 0.01247$ & $2.53500\pm 0.07367$ & $\best{0.05252}\pm 0.00061$ & $0.26289\pm 0.01034$ & $0.22930\pm 0.00830$ \\
 & $\checkmark$ & -- & $0.18923\pm 0.00666$ & $0.06722\pm 0.01546$ & $2.51722\pm 0.07022$ & $0.05448\pm 0.00265$ & $0.26407\pm 0.00727$ & $\best{0.22529}\pm 0.00982$ \\
 & -- & $\checkmark$ & $0.19125\pm 0.00990$ & $0.05942\pm 0.03401$ & $2.54883\pm 0.12060$ & $0.05974\pm 0.00794$ & $0.26551\pm 0.01164$ & $0.23400\pm 0.02137$ \\
 & $\checkmark$ & $\checkmark$ & $\best{0.18506}\pm 0.01290$ & $\best{0.05868}\pm 0.01903$ & $\best{2.47311}\pm 0.15687$ & $0.05575\pm 0.00157$ & $\best{0.25740}\pm 0.01184$ & $0.23088\pm 0.01935$ \\
\midrule
Weather & -- & -- & $0.22187\pm 0.00429$ & $0.04222\pm 0.00459$ & $1.51642\pm 0.01457$ & $0.08509\pm 0.00160$ & $0.28126\pm 0.00802$ & $0.24990\pm 0.00636$ \\
 & $\checkmark$ & -- & $0.22567\pm 0.00861$ & $0.04598\pm 0.00375$ & $1.54189\pm 0.04017$ & $0.08586\pm 0.00354$ & $0.28377\pm 0.01162$ & $0.25434\pm 0.01535$ \\
 & -- & $\checkmark$ & $0.22075\pm 0.00402$ & $0.04142\pm 0.00539$ & $1.51521\pm 0.02164$ & $0.08528\pm 0.00148$ & $\best{0.28026}\pm 0.00739$ & $0.24873\pm 0.00825$ \\
 & $\checkmark$ & $\checkmark$ & $\best{0.21513}\pm 0.00480$ & $\best{0.03108}\pm 0.00685$ & $\best{1.48730}\pm 0.02395$ & $\best{0.08364}\pm 0.00190$ & $0.28153\pm 0.00381$ & $\best{0.24823}\pm 0.00495$ \\
\bottomrule
\end{tabular*}
\end{table*}

\paragraph{Comparison with direct Gaussian forecasts}
We independently train a conditional Gaussian predictor with the same
center, variance, factor and correction architecture, but without the
denoising loss or reverse diffusion chain. It uses the same maximum stage
budgets and patience, and selects post-pretraining checkpoints by validation CRPS.
Table~\ref{tab:gaussian-comparison} compares this predictor with the
primary FaCoDiff benchmarks on Solar and Weather.
FaCoDiff has lower means on five of six metrics on Solar and all six on
Weather. CRPS and Energy decrease by 3.41\% and 2.80\% on Solar and by
11.91\% and 2.31\% on Weather. Weather Variogram also decreases by
10.08\%, while the direct Gaussian predictor retains the lower Variogram
on Solar.

\begin{table*}[tbp]
\centering
\rmfamily\fontsize{7.2}{8.6}\selectfont
\setlength{\tabcolsep}{3pt}
\renewcommand{\arraystretch}{0.96}
\caption{Independently trained Gaussian and diffusion forecasts on Solar and Weather. Results are mean $\pm$ standard deviation over three runs. Lower is better; red marks the better mean.}
\label{tab:gaussian-comparison}
\begin{tabular*}{\textwidth}{@{\extracolsep{\fill}}llcccccc@{}}
\toprule
Dataset & Variant & CRPS & QICE & Energy & Variogram & MAE & MSE \\
\midrule
Solar & Gaussian & $0.19159\pm 0.01271$ & $0.06189\pm 0.02107$ & $2.54430\pm 0.15398$ & $\best{0.05385}\pm 0.00280$ & $0.26571\pm 0.01579$ & $0.23170\pm 0.02114$ \\
 & FaCoDiff & $\best{0.18506}\pm 0.01290$ & $\best{0.05868}\pm 0.01903$ & $\best{2.47311}\pm 0.15687$ & $0.05575\pm 0.00157$ & $\best{0.25740}\pm 0.01184$ & $\best{0.23088}\pm 0.01935$ \\
\midrule
Weather & Gaussian & $0.24422\pm 0.00445$ & $0.04054\pm 0.00346$ & $1.52244\pm 0.02211$ & $0.09301\pm 0.00197$ & $0.29467\pm 0.00685$ & $0.26180\pm 0.00769$ \\
 & FaCoDiff & $\best{0.21513}\pm 0.00480$ & $\best{0.03108}\pm 0.00685$ & $\best{1.48730}\pm 0.02395$ & $\best{0.08364}\pm 0.00190$ & $\best{0.28153}\pm 0.00381$ & $\best{0.24823}\pm 0.00495$ \\
\bottomrule
\end{tabular*}
\end{table*}

\subsection{Synthetic experiments}
\label{sec:synthetic}
We next test whether conditional factors benefit joint forecasts when the
true dependence is controlled while the generating marginal means and
variances remain fixed. We generate noisy sinusoidal histories with an observable
binary state $c(X)$ and 32-dimensional futures of length 24 from histories
of length 48. At prediction step $h$, the data follow
\begin{equation}
\begin{aligned}
 y_h&=\mu_h^\star(X)+\sqrt{v_h^\star}
 \left[\sqrt{1-\rho_c}\,\epsilon_h+\sqrt{\rho_c}\,B_c z_h\right],\\
 \epsilon_h&\sim\mathcal N(0,I_{32}),\qquad z_h\sim\mathcal N(0,I_4),
\end{aligned}
\label{eq:synthetic-generation}
\end{equation}
with independent Gaussian draws. Each row of $B_c\in\R^{32\times4}$ has
unit norm, so the conditional marginal variance is $v_h^\star$ for every
variable. We set $\mu_h^\star(X)=e^{-h/24}X_N$ and
$v_h^\star=0.6+0.2\sin^2(2\pi h/24)$. The three scenarios are
\emph{Independent} ($\rho_c=0$), \emph{Fixed global} (a common $B_0$), and
\emph{State-switching} (two distinct rank-four loadings $B_0,B_1$).
In the correlated scenarios, $\rho_0=0.25$ and $\rho_1=0.60$.
Thus Fixed global fixes the loading basis, while its dependence strength
still varies with state. To construct the histories, let $M$ be the
order-32 Hadamard matrix and $\tau_n=(n-1)/48$. We generate
\begin{equation*}
\begin{aligned}
 X_{nj}&=\frac1{\sqrt8}\sum_{k=1}^8 M_{j,k+1}\sin(2\pi k\tau_n)\\
 &\quad+0.8q\cos(22\pi\tau_n)+0.15\xi_{nj},
\end{aligned}
\end{equation*}
where $q\in\{-1,1\}$ is equiprobable and $\xi_{nj}$ is independent
standard Gaussian noise. The state $c(X)$ is one when
$\sum_n(d^{-1}\sum_jX_{nj})\cos(22\pi\tau_n)>0$, and zero otherwise.
The bases $B_0$ and $B_1$ contain columns 2--5 and 6--9 of $M$,
respectively, rescaled by $1/2$.
The histories and generating marginal moments are matched across scenarios;
state labels and true factors are not supplied to the learned models.

Each seed provides 768 independent episodes, split into 537 training,
78 validation, and 153 test episodes. Standardization uses training episodes
only, separately within each scenario, and all variants use the same data.
For each scenario and seed, the three variants share a pretrained base checkpoint
and retain Location/Scale. Diagonal+L+S disables factor covariance; Global
factor learns a history-invariant shared basis with conditional coefficients
and strengths; Conditional factor uses the history-dependent basis of
\ours{}. All variants use capacity eight, fixed 20/10/20-epoch base/Factor/joint
budgets, validation-CRPS checkpoint selection, and 100 samples per test
episode. The global variant adds 256 basis parameters; capacity eight can
span both true rank-four subspaces. Each variant learns its own means
and scales during training.

Table~\ref{tab:synthetic-factor} reports all three scenarios and the six
forecasting metrics. We additionally apply empirical CRPS to the normalized sum
$a_h=d^{-1/2}\sum_j y_{hj}$ of training-standardized variables, denoting
this auxiliary score by $\mathrm{CRPS}_{\mathrm{agg}}$.

\begin{table}[tbp]
\centering\tablefont
\setlength{\tabcolsep}{5pt}
\caption{Synthetic forecasting under three dependence scenarios. Mean $\pm$ standard deviation over three runs; red marks the lowest mean. $\mathrm{CRPS}_{\mathrm{agg}}$ scores the normalized sum of training-standardized variables defined in the text. Lower is better.}
\label{tab:synthetic-factor}
\begin{tabular*}{\linewidth}{@{\extracolsep{\fill}}lccc@{}}
\toprule
Metric & Diagonal+L+S & Global factor & Conditional factor \\
\midrule
\multicolumn{4}{l}{\textbf{Independent}}\\
CRPS & $0.52671\pm 0.00009$ & $0.52829\pm 0.00007$ & $\best{0.52666}\pm 0.00060$ \\
QICE & $\best{0.00558}\pm 0.00055$ & $0.00614\pm 0.00048$ & $0.00559\pm 0.00052$ \\
Energy & $3.70529\pm 0.00063$ & $3.71685\pm 0.00035$ & $\best{3.70498}\pm 0.00387$ \\
Variogram & $0.16823\pm 0.00026$ & $0.16853\pm 0.00022$ & $\best{0.16823}\pm 0.00023$ \\
MAE & $0.73964\pm 0.00078$ & $0.74169\pm 0.00079$ & $\best{0.73958}\pm 0.00138$ \\
MSE & $0.86185\pm 0.00053$ & $0.86680\pm 0.00020$ & $\best{0.86171}\pm 0.00222$ \\
$\mathrm{CRPS}_{\mathrm{agg}}$ & $\best{0.53566}\pm 0.01302$ & $0.53578\pm 0.01166$ & $0.53619\pm 0.01308$ \\
\midrule
\multicolumn{4}{l}{\textbf{Fixed global}}\\
CRPS & $\best{0.52796}\pm 0.00078$ & $0.52861\pm 0.00021$ & $0.52832\pm 0.00043$ \\
QICE & $0.00566\pm 0.00035$ & $0.00576\pm 0.00098$ & $\best{0.00333}\pm 0.00265$ \\
Energy & $3.69767\pm 0.00712$ & $3.70193\pm 0.00223$ & $\best{3.69358}\pm 0.00925$ \\
Variogram & $0.17201\pm 0.00038$ & $0.17184\pm 0.00024$ & $\best{0.17051}\pm 0.00126$ \\
MAE & $\best{0.74116}\pm 0.00073$ & $0.74208\pm 0.00022$ & $0.74213\pm 0.00027$ \\
MSE & $\best{0.86607}\pm 0.00324$ & $0.86816\pm 0.00176$ & $0.86812\pm 0.00103$ \\
$\mathrm{CRPS}_{\mathrm{agg}}$ & $0.41710\pm 0.00495$ & $0.42662\pm 0.00437$ & $\best{0.41076}\pm 0.01045$ \\
\midrule
\multicolumn{4}{l}{\textbf{State-switching}}\\
CRPS & $\best{0.52847}\pm 0.00163$ & $0.52936\pm 0.00101$ & $0.52884\pm 0.00083$ \\
QICE & $0.00536\pm 0.00022$ & $0.00560\pm 0.00065$ & $\best{0.00256}\pm 0.00106$ \\
Energy & $3.70094\pm 0.01274$ & $3.70719\pm 0.00896$ & $\best{3.69163}\pm 0.00285$ \\
Variogram & $0.17148\pm 0.00072$ & $0.17143\pm 0.00075$ & $\best{0.16932}\pm 0.00118$ \\
MAE & $\best{0.74191}\pm 0.00192$ & $0.74322\pm 0.00081$ & $0.74312\pm 0.00074$ \\
MSE & $\best{0.86781}\pm 0.00553$ & $0.87057\pm 0.00361$ & $0.87012\pm 0.00333$ \\
$\mathrm{CRPS}_{\mathrm{agg}}$ & $0.41733\pm 0.00449$ & $0.42682\pm 0.01384$ & $\best{0.40629}\pm 0.01250$ \\
\bottomrule
\end{tabular*}
\end{table}

\paragraph{Joint and aggregate prediction}
Conditional factors improve both joint-score means and
$\mathrm{CRPS}_{\mathrm{agg}}$ in the two correlated scenarios. Relative
to Diagonal+L+S, they reduce Energy, Variogram and
$\mathrm{CRPS}_{\mathrm{agg}}$ by 0.11\%, 0.87\% and 1.52\% with a fixed
global basis, and by 0.25\%, 1.26\% and 2.65\% under state switching.
In the latter setting, aggregate CRPS is also 4.81\% below the Global
factor control, and both Variogram and aggregate CRPS improve over both
controls in all three runs. Alongside these joint and aggregate gains,
marginal CRPS, MAE and MSE remain within 0.27\% of the diagonal control,
with slightly higher means, while QICE decreases from 0.00536 to 0.00256.
Under independent noise, the conditional and diagonal models are nearly
indistinguishable in the joint scores, and aggregate CRPS differs by
0.10\%. The contrast links the additional benefit to shared uncertainty.

\begin{figure*}[pos=t]
\centering
\includegraphics[width=\textwidth]{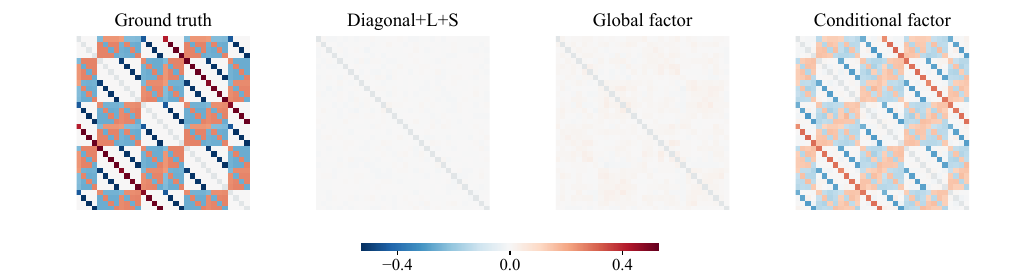}
\leftfigurecaption
\caption{True and generated forecast covariances in the stronger-dependence state of the state-switching scenario. Forecast matrices are averaged over three runs; all panels share a color scale, with gray diagonals.}
\label{fig:synthetic-covariance}
\end{figure*}

\begin{table}[tbp]
\centering\tablefont
\setlength{\tabcolsep}{5pt}
\caption{Off-diagonal covariance RMSE of final generated forecasts, averaged over both states within each run. Mean $\pm$ standard deviation over three runs; red marks the lowest mean. Lower is better.}
\label{tab:synthetic-covariance-error}
\begin{tabular*}{\linewidth}{@{\extracolsep{\fill}}lccc@{}}
\toprule
Scenario & Diagonal+L+S & Global factor & Conditional factor \\
\midrule
Independent & $\best{0.00738}\pm 0.00006$ & $0.00782\pm 0.00008$ & $0.00756\pm 0.00009$ \\
Fixed global & $0.17036\pm 0.00051$ & $0.16770\pm 0.00397$ & $\best{0.14888}\pm 0.01717$ \\
State-switching & $0.17034\pm 0.00045$ & $0.16894\pm 0.00235$ & $\best{0.10471}\pm 0.05002$ \\
\bottomrule
\end{tabular*}
\end{table}

\paragraph{Covariance in the generated forecasts}
Figure~\ref{fig:synthetic-covariance} and
Table~\ref{tab:synthetic-covariance-error} examine dependence in the final
diffusion forecasts. We use the first eight test episodes in each
state and estimate covariance across 100 generated samples separately at
each episode and horizon, then average over horizons and episodes. This
within-forecast centering excludes variation in conditional means. The
figure shows the three-run mean matrices for the stronger-dependence
state of the state-switching scenario. The table reports off-diagonal
RMSE averaged across states within each run, followed by the mean and
standard deviation across three runs.
In the state-switching scenario, the resulting error is 0.10471 for
Conditional factor, versus 0.17034 for Diagonal+L+S and 0.16894 for Global
factor, a 38.5\% reduction relative to the diagonal control. Recovery
strength varies across runs, but the lower mean error accompanies better
joint and aggregate scores. Thus the learned structure is reflected in
the final forecast ensemble, beyond the auxiliary Gaussian fit.

\subsection{Sensitivity to factor capacity}
\label{sec:rank-study}
We test how much factor capacity is needed by separately training
$r\in\{1,2,4,8,16\}$ on Solar and Weather.
Each capacity uses seeds 1/2/3, the same base within each seed, and matched
continuation budgets. Capacity changes both covariance and correction heads;
it is a sensitivity study of the complete factor estimator.

\begin{figure*}[pos=tbp]
\centering
\includegraphics[width=\textwidth]{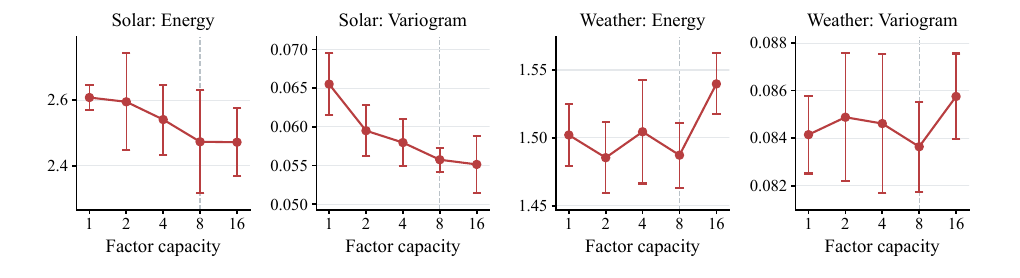}
\caption{Factor-capacity sensitivity on Solar and Weather. Points and error bars show three-run means and standard deviations; dashed lines mark the default capacity eight. Lower is better.}
\label{fig:capacity}
\end{figure*}

Figure~\ref{fig:capacity} shows that compact factor spaces capture most of
the gains available in this sweep. On Solar, increasing capacity from one
to eight reduces Energy and Variogram by 5.17\% and 14.91\%, respectively.
Doubling capacity from eight to sixteen yields a further 0.05\% and
1.09\% reduction in these scores. On Weather, capacities two and eight
achieve the lowest Energy and Variogram means, respectively; increasing
capacity to sixteen yields higher means on both scores. These results
support eight as an effective common setting.

\subsection{Effect of cross-variable dependence}
\label{sec:pairing}
To assess the predictive value of cross-variable pairing, we independently
permute sample indices for each variable at each test origin, holding
each variable's permutation fixed over the forecast horizon. This changes
how variable-specific sample trajectories are paired while preserving
every sampled univariate trajectory and marginal sample set.

\begin{figure*}[pos=tbp]
\centering
\includegraphics[width=\textwidth]{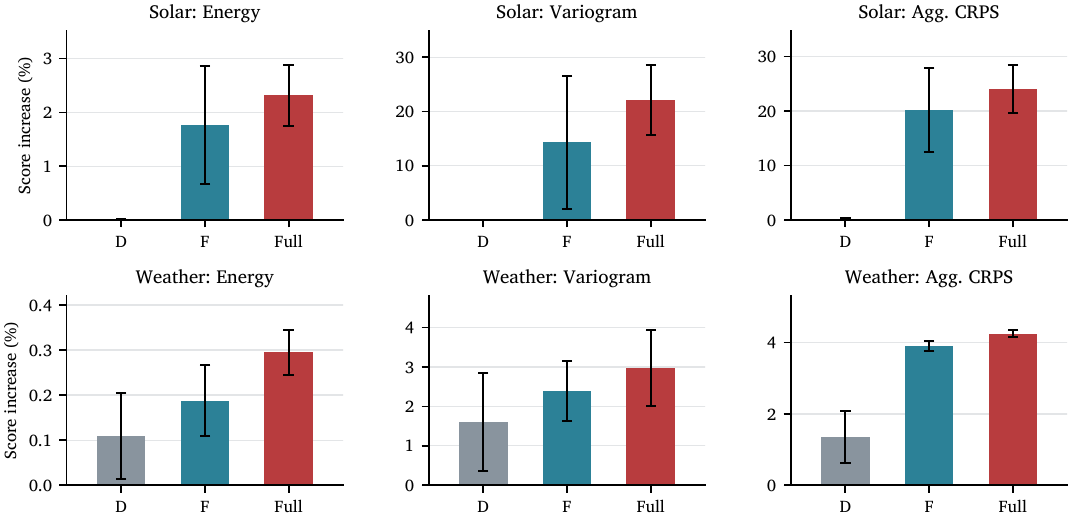}
\caption{Cross-variable sample-pairing permutation preserving univariate trajectories on Solar and Weather. Bars and error bars show three-run means and standard deviations of relative score increases; positive values indicate degradation. All panels compare D, F and Full; aggregate CRPS uses Eq.~\eqref{eq:aggregate-score}. D, F and Full follow Table~\ref{tab:ablation-components}.}
\label{fig:pairing}
\end{figure*}

Permutation worsens Full's Energy, Variogram and aggregate CRPS on both
datasets (Figure~\ref{fig:pairing}). On Solar, Variogram and aggregate
CRPS increase by 22.11\% and 24.06\%, respectively; on Weather, the
corresponding increases are 2.97\% and 4.25\%, averaging the three per-run
relative changes. The mean increases for F and Full exceed those for D
on all three scores in both datasets, linking the effect to factor
modeling. Since the marginal forecasts are unchanged, this deterioration
shows that cross-variable pairing itself has predictive value for joint
and aggregate forecasts, complementing the matched covariance controls
in Table~\ref{tab:ablation-components}.

\subsection{Inference efficiency}
\label{sec:efficiency}
On NVIDIA RTX4090 GPUs, we generate 100 samples at eight fixed, evenly spaced
complete test origins. After two warmup calls, five CUDA-synchronized repeats
measure the predictors, corrections, covariance preparation, reverse loop
and output transfer, excluding model loading and metric computation.
Table~\ref{tab:efficiency} reports mean time and sample standard deviation
per origin over the five repeats, and peak allocated CUDA memory.
CW-NsDiff uses its high-dimensional accelerated implementation.
The diagonal cost control retains the predictors, correction architecture
and factor computation but disables factor covariance during inference.

\begin{table}[tbp]
\centering\tablefont
\setlength{\tabcolsep}{4pt}
\caption{Generation time and peak allocated CUDA memory on RTX4090 GPUs. Time is mean $\pm$ standard deviation over five repeats, per origin for 100 samples.}
\label{tab:efficiency}
\begin{tabular*}{\linewidth}{@{}l@{\extracolsep{\fill}}lcc@{}}
\toprule
Dataset & Model & Time (s) & MiB \\
\midrule
Solar & NsDiff & $0.3080\pm0.0022$ & 132.5 \\
 & CW-NsDiff & $0.2920\pm0.0016$ & 241.6 \\
 & Diagonal+L+S & $0.2328\pm0.0079$ & 141.2 \\
 & FaCoDiff & $0.2314\pm0.0044$ & 141.3 \\
\midrule
ECL & NsDiff & $0.3309\pm0.0076$ & 214.4 \\
 & CW-NsDiff & $0.3091\pm0.0046$ & 1679.3 \\
 & Diagonal+L+S & $0.2511\pm0.0045$ & 224.7 \\
 & FaCoDiff & $0.2423\pm0.0094$ & 225.6 \\
\midrule
Taxi & NsDiff & $0.9048\pm0.1020$ & 156.7 \\
 & CW-NsDiff\textsuperscript{\(\dagger\)} & $0.4244\pm0.0044$ & 17375.8 \\
 & Diagonal+L+S & $0.2327\pm0.0041$ & 163.8 \\
 & FaCoDiff & $0.2353\pm0.0060$ & 163.9 \\
\bottomrule
\end{tabular*}
\par\smallskip
\begin{minipage}{\linewidth}
\footnotesize\textsuperscript{\(\dagger\)}\,Randomly initialized model used to measure architecture costs.
\end{minipage}
\end{table}

\paragraph{Generation time and memory}
FaCoDiff generates 100 samples per origin in 0.2314, 0.2423 and 0.2353\,s
on Solar, ECL and Taxi, respectively, with lower mean times than both
reported baseline implementations. Its peak allocated memory is 141.3,
225.6 and 163.9\,MiB. Relative to the matched diagonal inference-cost
control, the additional allocation is below 1\,MiB in all three settings,
and mean generation times differ by at most 3.5\%. Thus correlated
generation remains close in cost to the same conditional architecture
with diagonal noise. These are full-generation implementation costs;
covariance-operation complexity is analyzed separately in
Section~\ref{sec:lowrank}.

\paragraph{Scaling with dimension}
We measure generation cost at $d\in\{32,64,128,256,512,862,1000\}$ using
common synthetic inputs and randomly initialized models. History and forecast lengths are fixed
at 168 and 192, with 20 diffusion steps, 100 samples in chunks of ten,
and factor capacity eight. We use the same GPU model and timing protocol
as above, keeping CW-NsDiff's accelerated implementation fixed across
dimensions. The diagonal control retains the correction architecture
and factor computation but disables factor covariance during generation.

\begin{figure*}[pos=tbp]
\centering
\includegraphics[width=0.88\textwidth]{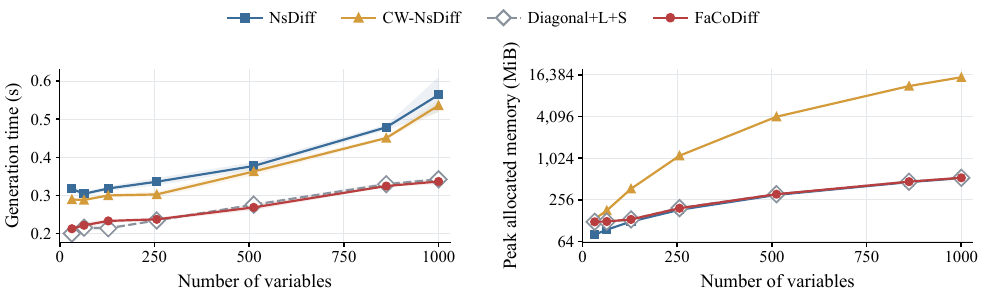}
\caption{Controlled architecture-cost scaling with randomly initialized models. Left: generation time for 100 samples per origin; bands show standard deviations over five repeats. Right: peak allocated CUDA memory on a logarithmic axis.}
\label{fig:dimension_scaling}
\end{figure*}

Figure~\ref{fig:dimension_scaling} shows that FaCoDiff remains close to
its diagonal control as dimension increases, while CW-NsDiff's memory
allocation grows much more rapidly. At $d=1000$, FaCoDiff requires
535.2\,MiB and 0.3366\,s per origin, compared with 15{,}168.5\,MiB and
0.5363\,s for CW-NsDiff: the dense implementation allocates 28.3 times as
much memory. This controlled profile demonstrates the practical memory
advantage of structured generation at larger dimensions.
\FloatBarrier

\section{Conclusion}
\label{sec:conclusion}
We presented \ours{} as a way to incorporate conditional cross-variable
dependence into diffusion forecasting while retaining control over marginal
noise scales. Its most consistent empirical strength is dependence-sensitive
forecast quality, with the lowest mean Variogram score on eight of nine
datasets. The matched controls and sample-pairing interventions connect this
advantage to dependence that improves joint and aggregate forecasts. Together,
these results show the value of treating conditional dependence as an explicit
part of the statistical reference and generation process.

The low-rank construction makes this approach practical for higher-dimensional
forecasting, with generation costs close to those of its diagonal control in
the reported measurements. For researchers developing probabilistic
forecasters, the residual-based estimator and matrix-free operators provide a
concrete route to introducing structured noise into conditional generators
while retaining tractable likelihood evaluation and noise transport. The
pairing intervention also provides a useful diagnostic for assessing whether
generated dependence contributes predictive information beyond the marginal
sample sets.

The current model treats forecast steps as conditionally independent and uses
a fixed factor capacity across the main benchmarks. Future work can introduce
temporal coupling into the factor geometry and select factor capacity
according to the strength and complexity of residual co-movement. Another
direction is to examine how statistical-reference accuracy and reverse-chain
length jointly affect forecast quality, clarifying how residual estimation and
finite-step denoising should be coordinated.

\section*{CRediT authorship contribution statement}
\textbf{Jiaxuan Li:} Conceptualization, Methodology, Software, Validation,
Formal analysis, Investigation, Resources, Data curation, Visualization,
Supervision, Project administration, Writing --
original draft, Writing -- review \& editing.
\textbf{Yao Liu:} Supervision, Funding acquisition, Writing -- review \& editing.
\textbf{Tingting Dai:} Supervision, Funding acquisition, Writing -- review \& editing.
\textbf{Yuewei Xue:} Conceptualization, Writing -- review \& editing.
\textbf{Shijie Dong:} Conceptualization, Writing -- review \& editing.
\textbf{Zixiang Jin:} Conceptualization, Writing -- review \& editing.
\textbf{Xiaoyu Guo:} Conceptualization, Writing -- review \& editing.
\textbf{Qiao Liu:} Supervision, Funding acquisition, Writing -- review \& editing.

\section*{Declaration of competing interest}
The authors declare that they have no known competing financial interests
or personal relationships that could have appeared to influence the work
reported in this paper.

\section*{Data and code availability}
The benchmark datasets are publicly available from the sources cited in
Section~\ref{sec:experiments}. The source code will be made publicly
available upon acceptance.

\section*{Declaration of generative AI and AI-assisted technologies in the manuscript preparation process}
During the preparation of this work, the authors used ChatGPT/Codex for
language polishing and LaTeX formatting. The authors reviewed and edited
the output as needed and take full responsibility for the content of the
published article.

\section*{Funding}
This work was supported by the National Natural Science Foundation of China \mbox{(U22B2061)},
the National Key R\&D Program of China \mbox{(2022YFB4300603)},
and the Natural Science Foundation of Sichuan, China \mbox{(2026NSFSC0429)}.

\bibliographystyle{elsarticle-num}
\bibliography{references}

\begin{thebibliography}{10}
\expandafter\ifx\csname url\endcsname\relax
  \def\url#1{\texttt{#1}}\fi
\expandafter\ifx\csname urlprefix\endcsname\relax\def\urlprefix{URL }\fi
\expandafter\ifx\csname href\endcsname\relax
  \def\href#1#2{#2} \def\path#1{#1}\fi

\bibitem{gneiting2007scoring}
T.~Gneiting, A.~E. Raftery, Strictly proper scoring rules, prediction, and
  estimation, J. Am. Stat. Assoc. 102~(477) (2007) 359--378.
\newblock \href {https://doi.org/10.1198/016214506000001437}
  {\path{doi:10.1198/016214506000001437}}.

\bibitem{scheuerer2015variogram}
M.~Scheuerer, T.~M. Hamill, Variogram-based proper scoring rules for
  probabilistic forecasts of multivariate quantities, Mon. Weather Rev. 143~(4)
  (2015) 1321--1334.
\newblock \href {https://doi.org/10.1175/MWR-D-14-00269.1}
  {\path{doi:10.1175/MWR-D-14-00269.1}}.

\bibitem{ho2020ddpm}
J.~Ho, A.~Jain, P.~Abbeel,
  \href{https://proceedings.neurips.cc/paper/2020/hash/4c5bcfec8584af0d967f1ab10179ca4b-Abstract.html}{Denoising
  diffusion probabilistic models}, in: Adv. Neural Inf. Process. Syst.,
  Vol.~33, 2020, pp. 6840--6851.
\newline\urlprefix\url{https://proceedings.neurips.cc/paper/2020/hash/4c5bcfec8584af0d967f1ab10179ca4b-Abstract.html}

\bibitem{ye2025nsdiff}
W.~Ye, Z.~Xu, N.~Gui,
  \href{https://proceedings.mlr.press/v267/ye25i.html}{Non-stationary diffusion
  for probabilistic time series forecasting}, in: Proc. Int. Conf. Mach.
  Learn., Vol. 267 of Proc. Mach. Learn. Res., PMLR, 2025, pp. 72112--72130.
\newline\urlprefix\url{https://proceedings.mlr.press/v267/ye25i.html}

\bibitem{lee2022priorgrad}
S.-g. Lee, H.~Kim, C.~Shin, X.~Tan, C.~Liu, Q.~Meng, T.~Qin, W.~Chen, S.~Yoon,
  T.-Y. Liu, \href{https://openreview.net/forum?id=_BNiN4IjC5}{{PriorGrad}:
  Improving conditional denoising diffusion models with data-dependent adaptive
  prior}, in: Proc. Int. Conf. Learn. Represent., 2022.
\newline\urlprefix\url{https://openreview.net/forum?id=_BNiN4IjC5}

\bibitem{yang2026cwgen}
Y.~Yang, S.~Chen, P.~Hu, Z.~Shen, Y.~Zhang, Z.~Sun, S.~Li, Z.~Chen,
  K.~Fukumizu,
  \href{https://proceedings.iclr.cc/paper_files/paper/2026/hash/57587d8d6a7ede0e5302fc22d0878c53-Abstract-Conference.html}{Conditionally
  whitened generative models for probabilistic time series forecasting}, in:
  Proc. Int. Conf. Learn. Represent., 2026.
\newline\urlprefix\url{https://proceedings.iclr.cc/paper_files/paper/2026/hash/57587d8d6a7ede0e5302fc22d0878c53-Abstract-Conference.html}

\bibitem{salinas2019copula}
D.~Salinas, M.~Bohlke-Schneider, L.~Callot, R.~Medico, J.~Gasthaus,
  \href{https://proceedings.neurips.cc/paper/2019/hash/0b105cf1504c4e241fcc6d519ea962fb-Abstract.html}{High-dimensional
  multivariate forecasting with low-rank gaussian copula processes}, in: Adv.
  Neural Inf. Process. Syst., Vol.~32, 2019.
\newline\urlprefix\url{https://proceedings.neurips.cc/paper/2019/hash/0b105cf1504c4e241fcc6d519ea962fb-Abstract.html}

\bibitem{zheng2024correlated}
V.~Z. Zheng, L.~Sun, Multivariate probabilistic time series forecasting with
  correlated errors, in: Adv. Neural Inf. Process. Syst., Vol.~37, 2024, pp.
  54288--54329.
\newblock \href {https://doi.org/10.52202/079017-1720}
  {\path{doi:10.52202/079017-1720}}.

\bibitem{murray2013copula}
J.~S. Murray, D.~B. Dunson, L.~Carin, J.~E. Lucas, Bayesian gaussian copula
  factor models for mixed data, J. Am. Stat. Assoc. 108~(502) (2013) 656--665.
\newblock \href {https://doi.org/10.1080/01621459.2012.762328}
  {\path{doi:10.1080/01621459.2012.762328}}.

\bibitem{lim2021survey}
B.~Lim, S.~Zohren, Time-series forecasting with deep learning: a survey,
  Philos. Trans. R. Soc. A 379~(2194) (2021) 20200209.
\newblock \href {https://doi.org/10.1098/rsta.2020.0209}
  {\path{doi:10.1098/rsta.2020.0209}}.

\bibitem{salinas2020deepar}
D.~Salinas, V.~Flunkert, J.~Gasthaus, T.~Januschowski, {DeepAR}: Probabilistic
  forecasting with autoregressive recurrent networks, Int. J. Forecast. 36~(3)
  (2020) 1181--1191.
\newblock \href {https://doi.org/10.1016/j.ijforecast.2019.07.001}
  {\path{doi:10.1016/j.ijforecast.2019.07.001}}.

\bibitem{rangapuram2018deepstate}
S.~S. Rangapuram, M.~W. Seeger, J.~Gasthaus, L.~Stella, Y.~Wang,
  T.~Januschowski,
  \href{https://papers.neurips.cc/paper_files/paper/2018/hash/5cf68969fb67aa6082363a6d4e6468e2-Abstract.html}{Deep
  state space models for time series forecasting}, in: Adv. Neural Inf.
  Process. Syst., Vol.~31, 2018.
\newline\urlprefix\url{https://papers.neurips.cc/paper_files/paper/2018/hash/5cf68969fb67aa6082363a6d4e6468e2-Abstract.html}

\bibitem{koenker1978quantiles}
R.~Koenker, G.~Bassett, Regression quantiles, Econometrica 46~(1) (1978)
  33--50.
\newblock \href {https://doi.org/10.2307/1913643} {\path{doi:10.2307/1913643}}.

\bibitem{wen2017mqrnn}
R.~Wen, K.~Torkkola, B.~Narayanaswamy, D.~Madeka,
  \href{https://www.amazon.science/publications/a-multi-horizon-quantile-recurrent-forecaster}{A
  multi-horizon quantile recurrent forecaster}, in: NIPS Time Series Workshop,
  2017.
\newline\urlprefix\url{https://www.amazon.science/publications/a-multi-horizon-quantile-recurrent-forecaster}

\bibitem{lim2021tft}
B.~Lim, S.~O. Arik, N.~Loeff, T.~Pfister, {Temporal Fusion Transformers} for
  interpretable multi-horizon time series forecasting, Int. J. Forecast. 37~(4)
  (2021) 1748--1764.
\newblock \href {https://doi.org/10.1016/j.ijforecast.2021.03.012}
  {\path{doi:10.1016/j.ijforecast.2021.03.012}}.

\bibitem{wang2019deepfactors}
Y.~Wang, A.~Smola, D.~Maddix, J.~Gasthaus, D.~Foster, T.~Januschowski,
  \href{https://proceedings.mlr.press/v97/wang19k.html}{Deep factors for
  forecasting}, in: Proc. Int. Conf. Mach. Learn., Vol.~97, 2019, pp.
  6607--6617.
\newline\urlprefix\url{https://proceedings.mlr.press/v97/wang19k.html}

\bibitem{dinh2017realnvp}
L.~Dinh, J.~Sohl-Dickstein, S.~Bengio,
  \href{https://openreview.net/forum?id=HkpbnH9lx}{Density estimation using
  {Real NVP}}, in: Proc. Int. Conf. Learn. Represent., 2017.
\newline\urlprefix\url{https://openreview.net/forum?id=HkpbnH9lx}

\bibitem{rasul2021flows}
K.~Rasul, A.-S. Sheikh, I.~Schuster, U.~Bergmann, R.~Vollgraf,
  \href{https://openreview.net/forum?id=WiGQBFuVRv}{Multivariate probabilistic
  time series forecasting via conditioned normalizing flows}, in: Proc. Int.
  Conf. Learn. Represent., 2021.
\newline\urlprefix\url{https://openreview.net/forum?id=WiGQBFuVRv}

\bibitem{sohldickstein2015diffusion}
J.~Sohl-Dickstein, E.~Weiss, N.~Maheswaranathan, S.~Ganguli,
  \href{https://proceedings.mlr.press/v37/sohl-dickstein15.html}{Deep
  unsupervised learning using nonequilibrium thermodynamics}, in: Proc. Int.
  Conf. Mach. Learn., Vol.~37, 2015, pp. 2256--2265.
\newline\urlprefix\url{https://proceedings.mlr.press/v37/sohl-dickstein15.html}

\bibitem{song2021sde}
Y.~Song, J.~Sohl-Dickstein, D.~P. Kingma, A.~Kumar, S.~Ermon, B.~Poole,
  \href{https://openreview.net/forum?id=PxTIG12RRHS}{Score-based generative
  modeling through stochastic differential equations}, in: Proc. Int. Conf.
  Learn. Represent., 2021.
\newline\urlprefix\url{https://openreview.net/forum?id=PxTIG12RRHS}

\bibitem{chang2026fundamentals}
Z.~Chang, G.~A. Koulieris, H.~J. Chang, H.~P.~H. Shum, On the design
  fundamentals of diffusion models: A survey, Pattern Recognit. 169 (2026)
  111934.
\newblock \href {https://doi.org/10.1016/j.patcog.2025.111934}
  {\path{doi:10.1016/j.patcog.2025.111934}}.

\bibitem{rasul2021timegrad}
K.~Rasul, C.~Seward, I.~Schuster, R.~Vollgraf,
  \href{https://proceedings.mlr.press/v139/rasul21a.html}{Autoregressive
  denoising diffusion models for multivariate probabilistic time series
  forecasting}, in: Proc. Int. Conf. Mach. Learn., Vol. 139, 2021, pp.
  8857--8868.
\newline\urlprefix\url{https://proceedings.mlr.press/v139/rasul21a.html}

\bibitem{tashiro2021csdi}
Y.~Tashiro, J.~Song, Y.~Song, S.~Ermon,
  \href{https://proceedings.neurips.cc/paper/2021/hash/cfe8504bda37b575c70ee1a8276f3486-Abstract.html}{{CSDI}:
  Conditional score-based diffusion models for probabilistic time series
  imputation}, in: Adv. Neural Inf. Process. Syst., Vol.~34, 2021, pp.
  24804--24816.
\newline\urlprefix\url{https://proceedings.neurips.cc/paper/2021/hash/cfe8504bda37b575c70ee1a8276f3486-Abstract.html}

\bibitem{shen2023timediff}
L.~Shen, J.~Kwok,
  \href{https://proceedings.mlr.press/v202/shen23d.html}{Non-autoregressive
  conditional diffusion models for time series prediction}, in: Proc. Int.
  Conf. Mach. Learn., Vol. 202, 2023, pp. 31016--31029.
\newline\urlprefix\url{https://proceedings.mlr.press/v202/shen23d.html}

\bibitem{kollovieh2023tsdiff}
M.~Kollovieh, A.~F. Ansari, M.~Bohlke-Schneider, J.~Zschiegner, H.~Wang,
  Y.~Wang, Predict, refine, synthesize: Self-guiding diffusion models for
  probabilistic time series forecasting, in: Adv. Neural Inf. Process. Syst.,
  Vol.~36, 2023, pp. 28341--28364.
\newblock \href {https://doi.org/10.52202/075280-1232}
  {\path{doi:10.52202/075280-1232}}.

\bibitem{yang2027gpd4ts}
J.~Yang, T.~Dai, P.~Liu, J.~Wu, N.~Li, J.~Bao, H.~Zhang, S.~Xia, {GPD4TS}: A
  generative pretrained diffusion paradigm for time series forecasting, Pattern
  Recognit. 183 (2027) 114817.
\newblock \href {https://doi.org/10.1016/j.patcog.2026.114817}
  {\path{doi:10.1016/j.patcog.2026.114817}}.

\bibitem{han2022card}
X.~Han, H.~Zheng, M.~Zhou, {CARD}: Classification and regression diffusion
  models, in: Adv. Neural Inf. Process. Syst., Vol.~35, 2022, pp. 18100--18115.
\newblock \href {https://doi.org/10.52202/068431-1316}
  {\path{doi:10.52202/068431-1316}}.

\bibitem{li2024tmdm}
Y.~Li, W.~Chen, X.~Hu, B.~Chen, B.~Sun, M.~Zhou,
  \href{https://proceedings.iclr.cc/paper_files/paper/2024/hash/516a9317af9d89e9f2251bd7fde49b8f-Abstract-Conference.html}{Transformer-modulated
  diffusion models for probabilistic multivariate time series forecasting}, in:
  Proc. Int. Conf. Learn. Represent., 2024.
\newline\urlprefix\url{https://proceedings.iclr.cc/paper_files/paper/2024/hash/516a9317af9d89e9f2251bd7fde49b8f-Abstract-Conference.html}

\bibitem{ye2026diffpts}
W.~Ye, D.~Li, H.~Liu, H.~Jiang, Y.~Sekimoto, R.~Jiang,
  \href{https://openreview.net/forum?id=x7OutBc7Vk}{{DiffPTS}: Rethinking
  diffusion {ELBO} for probabilistic time series forecasting}, in: Adv. Neural
  Inf. Process. Syst., 2026, accepted for NeurIPS 2026.
\newline\urlprefix\url{https://openreview.net/forum?id=x7OutBc7Vk}

\bibitem{rishi2025cndiff}
J.~Rishi, G.~Mothish, D.~Subramani,
  \href{https://proceedings.mlr.press/v267/rishi25a.html}{Conditional diffusion
  model with nonlinear data transformation for time series forecasting}, in:
  Proc. Int. Conf. Mach. Learn., Vol. 267, 2025, pp. 51703--51723.
\newline\urlprefix\url{https://proceedings.mlr.press/v267/rishi25a.html}

\bibitem{kim2022revin}
T.~Kim, J.~Kim, Y.~Tae, C.~Park, J.-H. Choi, J.~Choo,
  \href{https://openreview.net/forum?id=cGDAkQo1C0p}{Reversible instance
  normalization for accurate time-series forecasting against distribution
  shift}, in: Proc. Int. Conf. Learn. Represent., 2022.
\newline\urlprefix\url{https://openreview.net/forum?id=cGDAkQo1C0p}

\bibitem{liu2022nonstationary}
Y.~Liu, H.~Wu, J.~Wang, M.~Long, Non-stationary transformers: Exploring the
  stationarity in time series forecasting, in: Adv. Neural Inf. Process. Syst.,
  Vol.~35, 2022, pp. 9881--9893.
\newblock \href {https://doi.org/10.52202/068431-0718}
  {\path{doi:10.52202/068431-0718}}.

\bibitem{liu2024itransformer}
Y.~Liu, T.~Hu, H.~Zhang, H.~Wu, S.~Wang, L.~Ma, M.~Long,
  \href{https://proceedings.iclr.cc/paper_files/paper/2024/hash/2ea18fdc667e0ef2ad82b2b4d65147ad-Abstract-Conference.html}{{iTransformer}:
  Inverted transformers are effective for time series forecasting}, in: Proc.
  Int. Conf. Learn. Represent., 2024.
\newline\urlprefix\url{https://proceedings.iclr.cc/paper_files/paper/2024/hash/2ea18fdc667e0ef2ad82b2b4d65147ad-Abstract-Conference.html}

\bibitem{hou2025pmedgn}
M.~Hou, Z.~Liu, G.~Sa, Y.~Wang, J.~Sun, Z.~Li, J.~Tan, Parallel multi-scale
  dynamic graph neural network for multivariate time series forecasting,
  Pattern Recognit. 158 (2025) 111037.
\newblock \href {https://doi.org/10.1016/j.patcog.2024.111037}
  {\path{doi:10.1016/j.patcog.2024.111037}}.

\bibitem{engle2002dcc}
R.~Engle, Dynamic conditional correlation: A simple class of multivariate
  generalized autoregressive conditional heteroskedasticity models, J. Bus.
  Econ. Stat. 20~(3) (2002) 339--350.
\newblock \href {https://doi.org/10.1198/073500102288618487}
  {\path{doi:10.1198/073500102288618487}}.

\bibitem{ledoit2004shrinkage}
O.~Ledoit, M.~Wolf, A well-conditioned estimator for large-dimensional
  covariance matrices, J. Multivar. Anal. 88~(2) (2004) 365--411.
\newblock \href {https://doi.org/10.1016/S0047-259X(03)00096-4}
  {\path{doi:10.1016/S0047-259X(03)00096-4}}.

\bibitem{golub2013matrix}
G.~H. Golub, C.~F. Van~Loan, Matrix Computations, 4th Edition, Johns Hopkins
  University Press, 2013.

\bibitem{song2021ddim}
J.~Song, C.~Meng, S.~Ermon,
  \href{https://openreview.net/forum?id=St1giarCHLP}{Denoising diffusion
  implicit models}, in: Proc. Int. Conf. Learn. Represent., 2021.
\newline\urlprefix\url{https://openreview.net/forum?id=St1giarCHLP}

\bibitem{lu2022dpmsolver}
C.~Lu, Y.~Zhou, F.~Bao, J.~Chen, C.~Li, J.~Zhu, {DPM-Solver}: A fast {ODE}
  solver for diffusion probabilistic model sampling in around 10 steps, in:
  Adv. Neural Inf. Process. Syst., Vol.~35, 2022, pp. 5775--5787.
\newblock \href {https://doi.org/10.52202/068431-0418}
  {\path{doi:10.52202/068431-0418}}.

\bibitem{lai2018lstnet}
G.~Lai, W.-C. Chang, Y.~Yang, H.~Liu, Modeling long- and short-term temporal
  patterns with deep neural networks, in: Proc. Int. ACM SIGIR Conf. Res. Dev.
  Inf. Retr., 2018, pp. 95--104.
\newblock \href {https://doi.org/10.1145/3209978.3210006}
  {\path{doi:10.1145/3209978.3210006}}.

\bibitem{wu2021autoformer}
H.~Wu, J.~Xu, J.~Wang, M.~Long,
  \href{https://proceedings.neurips.cc/paper_files/paper/2021/hash/bcc0d400288793e8bdcd7c19a8ac0c2b-Abstract.html}{{Autoformer}:
  Decomposition transformers with auto-correlation for long-term series
  forecasting}, in: Adv. Neural Inf. Process. Syst., Vol.~34, 2021, pp.
  22419--22430.
\newline\urlprefix\url{https://proceedings.neurips.cc/paper_files/paper/2021/hash/bcc0d400288793e8bdcd7c19a8ac0c2b-Abstract.html}

\bibitem{zhou2021informer}
H.~Zhou, S.~Zhang, J.~Peng, S.~Zhang, J.~Li, H.~Xiong, W.~Zhang, {Informer}:
  Beyond efficient transformer for long sequence time-series forecasting, in:
  Proc. AAAI Conf. Artif. Intell., Vol.~35, 2021, pp. 11106--11115.
\newblock \href {https://doi.org/10.1609/aaai.v35i12.17325}
  {\path{doi:10.1609/aaai.v35i12.17325}}.

\bibitem{godahewa2021monash}
R.~Godahewa, C.~Bergmeir, G.~I. Webb, R.~J. Hyndman, P.~Montero-Manso,
  \href{https://datasets-benchmarks-proceedings.neurips.cc/paper_files/paper/2021/hash/eddea82ad2755b24c4e168c5fc2ebd40-Abstract-round2.html}{Monash
  time series forecasting archive}, in: Proc. NeurIPS Datasets Benchmarks
  Track, Vol.~1, 2021.
\newline\urlprefix\url{https://datasets-benchmarks-proceedings.neurips.cc/paper_files/paper/2021/hash/eddea82ad2755b24c4e168c5fc2ebd40-Abstract-round2.html}

\bibitem{alexandrov2020gluonts}
A.~Alexandrov, K.~Benidis, M.~Bohlke-Schneider, V.~Flunkert, J.~Gasthaus,
  T.~Januschowski, D.~C. Maddix, S.~Rangapuram, D.~Salinas, J.~Schulz,
  L.~Stella, A.~C. T{\"u}rkmen, Y.~Wang,
  \href{https://jmlr.org/papers/v21/19-820.html}{{GluonTS}: Probabilistic and
  neural time series modeling in python}, J. Mach. Learn. Res. 21~(116) (2020)
  1--6.
\newline\urlprefix\url{https://jmlr.org/papers/v21/19-820.html}

\end{thebibliography}
\end{document}